\documentclass[10pt,twocolumn,letterpaper]{article}

\usepackage{wacv}      

\definecolor{wacvblue}{rgb}{0.21,0.49,0.74}
\usepackage[pagebackref,breaklinks,colorlinks,allcolors=wacvblue]{hyperref}

\usepackage{latexsym}
\usepackage{amssymb}
\usepackage{amsmath}
\usepackage{amsthm}
\usepackage{booktabs}
\usepackage{enumitem}
\usepackage{graphicx}
\usepackage{color}

\usepackage{pifont} 
\usepackage{colortbl} 
\usepackage{xcolor}  
\usepackage{multirow}

\usepackage{url}

\usepackage{xcolor} 

\newcommand{\revision}[1]{\textcolor{black}{#1}}

\newcommand{\cmark}{\checkmark} 
\newcommand{\xmark}{\ding{55}}   

\newcommand{\BibTeX}{B\kern-.05em{\sc i\kern-.025em b}\kern-.08em\TeX}

\def\wacvPaperID{1320} 
\def\confName{WACV}
\def\confYear{2026}

\title{A Little More Like This: Text-to-Image Retrieval with Vision-Language Models Using Relevance Feedback}

\author{Bulat Khaertdinov, Mirela Popa and Nava Tintarev\\
Maastricht University\\
Maastricht, Netherlands \\
{\tt\small \{b.khaertdinov, mirela.popa, n.tintarev\}@maastrichtuniversity.nl}
}

\begin{document}
\maketitle

\begin{abstract}

Large vision-language models (VLMs) enable intuitive visual search using natural language queries. However, improving their performance often requires fine-tuning and scaling to larger model variants. In this work, we propose a mechanism inspired by traditional text-based search to improve retrieval performance at inference time: relevance feedback. While relevance feedback can serve as an alternative to fine-tuning, its model-agnostic design also enables use with fine-tuned VLMs. Specifically, we introduce and evaluate four feedback strategies for VLM-based retrieval. First, we revise classical pseudo-relevance feedback (PRF), which refines query embeddings based on top-ranked results. To address its limitations, we propose generative relevance feedback (GRF), which uses synthetic captions for query refinement. Furthermore, we introduce an attentive feedback summarizer (AFS), a custom transformer-based model that integrates multimodal fine-grained features from relevant items. Finally, we simulate explicit feedback using ground-truth captions as an upper-bound baseline. Experiments on Flickr30k and COCO with the VLM backbones show that GRF, AFS, and explicit feedback improve retrieval performance by 3–5\% in MRR@5 for smaller VLMs, and 1–3\% for larger ones, compared to retrieval with no feedback. Moreover, AFS, similarly to explicit feedback, mitigates query drift and is more robust than GRF in iterative, multi-turn retrieval settings. Our findings demonstrate that relevance feedback can consistently enhance retrieval across VLMs and open up opportunities for interactive and adaptive visual search.
\end{abstract}


\section{Introduction}

Many AI systems perform search through large multimodal databases. 
While modern image-text models, such as CLIP~\cite{radford2021learning} and BLIP~\cite{li2022blip, li2023blip} have advanced multimodal retrieval, they require substantial computational resources for fine-tuning and inference, especially with larger model variants. Motivated by classical information retrieval theory, we explore an alternative approach that can be combined with any pre-trained or fine-tuned VLM: \textit{relevance feedback}. Instead of modifying the retrieval model, this paradigm refines query vectors using representations of relevant and irrelevant items at inference time~\cite{rocchio1971relevance, salton1990improving}. Item relevance can be provided directly by the user, a strategy known as \textit{explicit feedback}. Relevance signals can also be approximated through \textit{pseudo-relevance feedback (PRF)}, which assumes top-ranked results are relevant~\cite{ruthven2003survey}. Another paradigm, namely \textit{generative relevance feedback (GRF)}, was recently introduced for text-based search, using large language models (LLMs) to expand or rewrite queries~\cite{mackie2023generative}. Despite its potential in text-based retrieval, relevance feedback remains largely underexplored in multimodal settings.


\begin{figure}[t]
    \centering
    \includegraphics[width=0.99\linewidth]{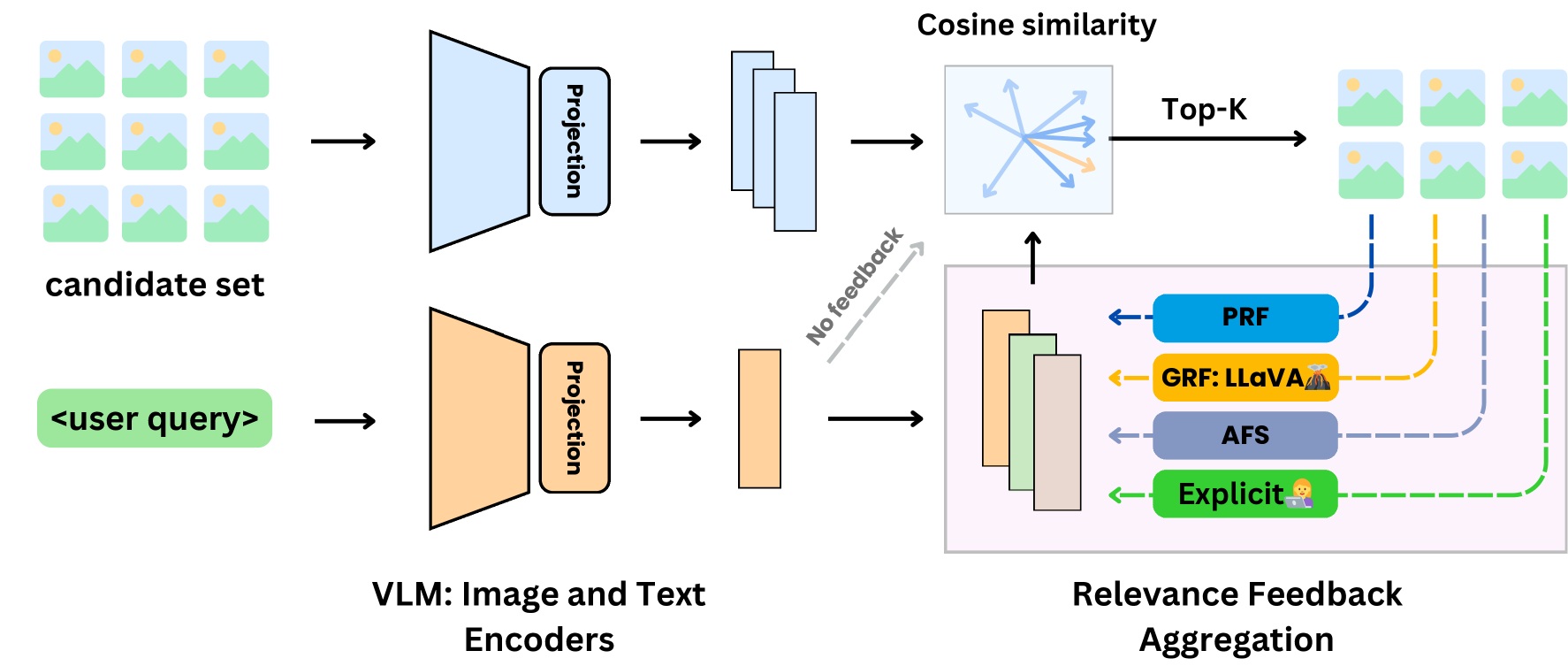}
    \caption{\textbf{Proposed VLM-based text-to-image retrieval with relevance feedback.} Query representations are refined in feature space using relevance feedback vectors.}
    \label{fig:retrieval_rocchio}
\end{figure}



This paper introduces a novel relevance feedback framework for VLM-based image retrieval that can operate with different types of feedback signals (Figure~\ref{fig:retrieval_rocchio}). Specifically, we propose and systematically evaluate four relevance feedback strategies: PRF, GRF, AFS, and simulated explicit user feedback. We demonstrate that relevance feedback can enhance retrieval performance at inference time. Crucially, our relevance feedback techniques are model-agnostic and can be applied on top of any image-text retrieval model, including fine-tuned VLMs. The main contributions of this work are as follows:

\begin{itemize}
    \item We extend classical Rocchio for PRF (Section~\ref{sec:meth_rocchio}) and introduce a conceptually novel GRF framework utilizing captions generated with LLaVA \cite{liu2024improved} (Section~\ref{sec:meth_grf}).
    \item We present a new attentive feedback summarizer (AFS) as a lightweight model that learns summarized feedback representations, which aggregate fine-grained signals at the patch and token levels (Section~\ref{sec:meth_afs}). Containing only two transformer blocks, AFS achieves performance often comparable to explicit relevance feedback. Additionally, attention visualization in AFS illustrates how feedback is aggregated from local image and caption parts.
    \item We evaluate four relevance feedback strategies against baseline retrieval (Section~\ref{sec:strategies}), on the Flickr30k \cite{young2014flickr} and COCO \cite{lin2014coco} datasets with four VLM models. Our findings highlight the shortcomings of PRF and the potential of GRF, AFS, and explicit feedback to enhance text-to-image retrieval at test time. Our codebase is available at: \url{https://github.com/bulatkh/vlm_retrieval_relevance_feedback}.
\end{itemize}

\section{Related Work}

Relevance feedback is a classical approach in document search that refines queries based on the relevance of retrieved documents. While widely used in text-based retrieval, its potential in visual applications remains underexplored. In this work, we adapt and extend relevance feedback for vision-language models, proposing novel strategies to enhance text-to-image retrieval.

\subsection{Relevance Feedback}
Classical text-based search relies on word frequency-based methods \cite{ramos2003using, robertson2009probabilistic, sparck1972statistical}, struggling with semantic ambiguity, synonymy, and polysemy \cite{deerwester1990indexing, krovetz1992lexical}. Furthermore, users may find it difficult to formulate effective queries due to their limited understanding of the collection and retrieval environment \cite{salton1990improving}. PRF techniques partially address these issues by refining query representations \cite{ruthven2003survey}. For instance, the Rocchio algorithm \cite{rocchio1971relevance} iteratively adjusts queries in vector space by reinforcing relevant document vectors and reducing the influence of non-relevant ones. 

Modern NLP models produce representations that reduce semantic ambiguity. Recent work has shown that pseudo-relevance feedback can improve dense retrieval by refining query and document embeddings~\cite{li2023pseudo, wang2021pseudo, yu2024tprf}. Other approaches combine relevance feedback with LLMs by reformulating the query in text space rather than modifying embeddings~\cite{jagerman2023query}, including GRF, where LLMs rewrite queries~\cite{mackie2023generative}. 

In this paper, we build on the principles of both PRF and GRF to develop novel relevance feedback methods tailored to vision-language models.

\subsection{Text-to-image Retrieval with VLMs}
Vision-language models, such as CLIP \cite{radford2021learning} and BLIP \cite{li2022blip, li2023blip}, have transformed visual search by mapping text and images into a shared feature space, enabling more effective retrieval. Despite being promising in text-based search with contemporary dense retrievers, relevance feedback adaptation in text-to-image retrieval is limited. Existing approaches are mostly based on a combination of explicit user feedback with LLMs applied at inference time. In 2024, \citet{lee2024interactive} and \citet{zhu2024enhancing} proposed LLM-based methods for query refinement using clarifying questions leveraging dialogue context and user feedback with synthetic image captions, respectively. \citeauthor{wang2024improving}~\cite{wang2024improving} utilized LLMs to score the relevance of images based on user interactions with image fragments. \revision{In 2025, \citeauthor{guofengding2025visual}~\cite{guofengding2025visual} proposed VISA, a test-time approach that uses representations of question answering-refined textual descriptions to re-rank the retrieved candidates.} These methods rely on prompting large generative models at inference time to enhance queries and summarize items retrieved with VLMs. In contrast, our work explores relevance feedback techniques that refine queries directly in the latent space of VLMs without requiring generative models at inference. 


\section{Methodology}

In text-to-image retrieval, the goal is to rank a set of $N$ candidate images $\{\boldsymbol{v}_i\}_{i=1}^N$ based on their relevance to textual query $\boldsymbol{q}$. We use a similarity-based retrieval framework with VLMs that project text (query) and image inputs into a shared latent space. These models produce global query and image representations, $\boldsymbol{z}_q, \boldsymbol{z}_i \in \mathbb{R}^d$, compared using cosine similarity to score and rank the candidates. Dimensionality $d$ is defined by the projection layer of the VLM.





\subsection{Relevance Feedback Aggregation with Rocchio}
\label{sec:meth_rocchio}

The classical Rocchio algorithm \cite{rocchio1971relevance} enables PRF by computing a refined representation of query $\boldsymbol{z}'_q \in \mathbb{R}^d$. This vector is obtained by fusing representations of the original query $\boldsymbol{z}_q$ with relevant and irrelevant (or positive and negative) feature vectors $\boldsymbol{z}_p$ and $\boldsymbol{z}_n$. In the original Rocchio, positive and negative vectors are obtained by averaging the most and least relevant items found by the retrieval model. We refer readers to App. \ref{sec:appendix_rocchio} for a detailed description of the original Rocchio formulation.\footnote{The appendices are shared through supplementary materials.} 

In this paper, we extend Rocchio for VLMs, suggesting two changes: (1) applying softmax-weighted averaging for feedback aggregation and (2) sourcing both positive and negative feedback from top-ranked items. Our weighing mechanism emphasizes representations of the most relevant candidates, assigning higher importance to feedback from candidates closer to the user query. Furthermore, by using top-ranked items for both positive and negative feedback, we ensure that the model focuses on hard negatives, i.e., the candidate items found relevant by a VLM that do not necessarily match the user query. Formally, we define the weight of candidate $i$ for a query $q$ as follows:

\begin{equation}
    w_{q, i} = \frac{exp(s_{q, i} / \tau)}{\sum_{c \in \boldsymbol{C_r}}exp(s_{q, c} / \tau)}
    \label{eq:weights}
\end{equation}

\noindent where $s_{q, i}$ is cosine similarity and $\tau$ is a temperature hyperparameter. Furthermore, positive and negative weights for feedback aggregation are defined as $w^p_{q, i} = w_{q, i}$ and $w^n_{q, i} = 1 - w_{q, i}$, respectively. The refined query can be written as follows:

\begin{equation}
    \boldsymbol{z}'_q = \alpha\boldsymbol{z}_q + \underbrace{\beta \sum_{i \in \boldsymbol{C}_r}w^p_{q, i}\boldsymbol{z}_i}_{\text{Positive relevance vector}} - \underbrace{\gamma\sum_{i \in \boldsymbol{C}_{r}}w^n_{q, i}\boldsymbol{z}_i}_{\text{Negative relevance vector}}
    \label{eq:rocchio_ours}
\end{equation}

\noindent where $\boldsymbol{C_r}$ is the set of top-ranked candidates. Therefore, relevance feedback is computed using top-K retrieved items, where $K = |\boldsymbol{C}_r|$. Temperature $\tau$ controls the sharpness of the weight distribution across items in $\boldsymbol{C}_r$. Lower values concentrate the weights, reducing overlap between positive and negative relevance vectors. We use $\tau = 0.05$ to sharpen the feedback signal, with $\alpha = 0.8$, $\beta = 0.1$, and $\gamma = 0.1$, as validated in Section~\ref{sec:results_ablations}.

\subsection{Generative Relevance Feedback}
\label{sec:meth_grf}
We introduce a novel version of generative relevance feedback (GRF) to enhance the query refinement process. Unlike GRF \cite{mackie2023generative} introduced for text-based search, we do not create multiple versions of the user query but rather caption the candidate images. Specifically, we leverage LLaVA-1.5 \cite{liu2024improved} to generate captions for all candidate images in a dataset. The synthetic captions can offer a rich feedback signal in the form of textual embeddings, capturing visual aspects that are not evident from the visual representations. 

To refine queries, we use our Rocchio algorithm (Section \ref{sec:meth_rocchio}) with relevance feedback from the generated captions. Namely, we aggregate representations of synthetic captions corresponding to retrieved images in line with Equation \ref{eq:rocchio_ours}. Unlike other methods that refine queries in natural text \cite{lee2024interactive, zhu2024enhancing}, our approach does not require using large generative AI models at inference time as synthetic captions for each candidate can be precomputed offline. 

\subsection{Attentive Feedback Summarizer}
\label{sec:meth_afs}
We identify several limitations of Rocchio when adapting it to text-to-image retrieval. First, the multimodality gap inherent in VLMs trained with contrastive objectives (e.g., CLIP, BLIP) could lead to suboptimal performance when combining visual and textual representations \cite{liang2022mind}. 
Furthermore, Rocchio is applied to aggregate global embeddings of images and captions produced by VLMs. Nevertheless, local information, e.g., fine-grained patch- and word-level representations, remains neglected, potentially limiting the ability to capture more nuanced relevance signals. Motivated by these challenges, we propose a compact model, denoted as attentive feedback summarizer~(AFS), that is trained to align feedback from different modalities with ground truth captions and images. 

 \begin{figure}[t]
    \centering
    \includegraphics[width=0.95\linewidth]{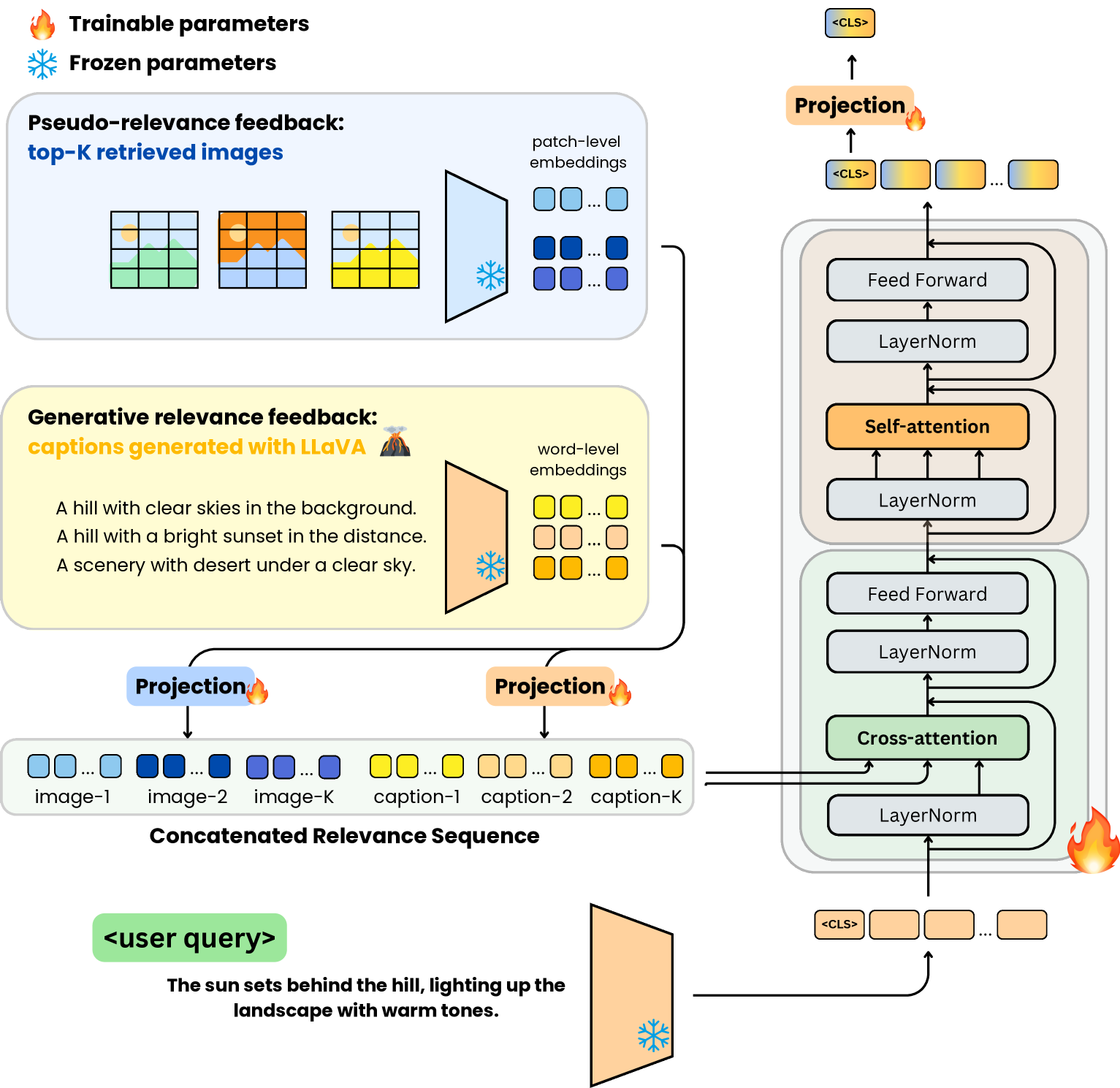}
    \caption{\textbf{Attentive Feedback Summarizer.} The user query, relevant images, and synthetic captions are processed using VLM encoders.}
    \label{fig:summarizer_architecture}
\end{figure}

\noindent\textbf{AFS Architecture:} 
 The attentive feedback summarizer (Figure \ref{fig:summarizer_architecture}) is a lightweight transformer model composed of two attention blocks \cite{vaswani2017attention}. First, fine-grained information from the relevance sequence is fused with query tokens through a cross-attention module. Self-attention queries are derived from the user query tokens. We also prepend a learnable \texttt{CLS} token similarly to BERT-like encoders \cite{devlin2019bert}. Keys and values are obtained from the relevance feedback, i.e., patch- and word-level representations extracted by the same VLM from the most relevant images and their synthetic captions. In this setup, the output of the cross-attention module consists of query representations contextualized by relevance feedback. Namely, each query token, along with the learnable \texttt{CLS} token, attends to local visual and textual features from the relevance sequence. The second component of the summarizer is a standard self-attention block that facilitates communication with the \texttt{CLS} token, which serves as the output of the feedback summarizer. A linear projection is applied to this token to obtain the final representation $\boldsymbol{z}^{\texttt{CLS}}_q \in \mathbb{R}^d$. A more thorough description of the AFS architecture can be found in App. \ref{sec:appendix_afs_att}.

\noindent\textbf{Training objective:} AFS is trained to produce a vector that summarizes relevance feedback, bridging the gap between the query and image embeddings. To achieve this, AFS learns representations aligned with the frozen representations of the ground-truth images and additional captions. We define cosine similarity loss\footnote{We opt for cosine similarity loss over contrastive losses because it allows training with smaller batches and reduces the number of hyperparameters. Since the VLM encoders are frozen, the representations are not expected to collapse, so techniques like negative pairs in contrastive learning or regularization are unnecessary.} for alignment with the ground-truth images and captions as follows:

\begin{equation}
    l^{\text{img}}_q = 1 - \frac{\boldsymbol{z}^{\text{CLS}}_q \cdot \boldsymbol{z}^\text{img}_q}{||\boldsymbol{z}^\text{CLS}_q||||\boldsymbol{z}^\text{img}_q||}
    \label{eq:img_loss}
\end{equation}

\begin{equation}
    l^{\text{cap}}_q = 1 - \frac{\boldsymbol{z}^\text{CLS}_q \cdot \boldsymbol{\overline{z}}^\text{cap}_q}{||\boldsymbol{z}^\text{CLS}_q||||\boldsymbol{\overline{z}}^\text{cap}_q||}
    \label{eq:cap_loss}
\end{equation}

\noindent where $\boldsymbol{z}^\text{img}_q \in \mathbb{R}^d$ is the ground truth image embedding, and $\boldsymbol{\overline{z}}^\text{cap}_q$ is the averaged representation of the additional captions in the dataset. Datasets used in this study provide multiple captions per image. For each training instance, we randomly sample one caption as a query, while the remaining ones are used for supervision along with the corresponding image. Finally, the accumulated loss for a mini-batch with $N_b$ queries is defined as follows: 

\begin{equation}
    \mathcal{L} = \frac{1}{2}\sum_{q=1}^{N_b}(l^\text{img}_q + l^\text{cap}_q)
    \label{eq:summarizer_objective}
\end{equation}

We investigate the impact of image and caption loss components through an ablation study presented in Section~\ref{sec:results_ablations}.


\noindent\textbf{Inference:} \revision{To incorporate AFS into the Rocchio framework, we define positive and negative vectors. We use the output of AFS $\boldsymbol{z}^{\texttt{CLS}}_q$ as the positive vector in Rocchio. 
However, AFS does not generate an output that provides negative relevance, as required by Rocchio. Learning the negative vector is a non-trivial task, as it would require access to irrelevant image and captions that can provide meaningful negative signals. Instead, we construct the negative relevance vector by aggregating the negated cross-attention scores. This enables the use of the AFS in Rocchio, even without providing negative feedback signals for training.} In detail, the cross-attention scores for query $q$ can be defined as $\boldsymbol{A}_q \in \mathbb{R}^{n_h \times s_q \times s_r}$, where $n_h$ is the number of attention heads, $s_q$ is the number of tokens in a user query, and $s_r = (s + p) * |\boldsymbol{C_r}|$ is the relevance sequence length, with $s$ and $p$ denoting the numbers of text tokens and image patch embeddings per relevant item, and $|\boldsymbol{C_r}|$ is the number of relevant items. We accumulate the cross-attention scores for each image and synthetic caption and obtain $\boldsymbol{A}^\text{img}_q = [A^\texttt{img}_{q, 1}, A^\texttt{img}_{q, 2}, \dots, A^\texttt{img}_{q, j},\dots, A^\texttt{img}_{q, |\boldsymbol{C}_r|}] \in \mathbb{R}^{|\boldsymbol{C}_r|}$ and, similarly, $\boldsymbol{A}^\text{cap}_q = [A^\texttt{cap}_{q, 1}, A^\texttt{cap}_{q, 2}, \dots, A^\texttt{cap}_{q, j},\dots, A^\texttt{cap}_{q, |\boldsymbol{C}_r|}] \in \mathbb{R}^{|\boldsymbol{C}_r|}$. Then, we compute weights for negative vector in Rocchio:

\begin{equation}
    w^\text{img}_{q, j} = \frac{exp((-A^\text{img}_{q,j}) / \tau)}{\sum_{c \in \boldsymbol{C_{r}}}exp((-A^\text{img}_{q, c}) / \tau)}
    \label{eq:img_neg_weight}
\end{equation}

\begin{equation}
    w^\text{cap}_{q, j} = \frac{exp((-A^\text{cap}_{q,j}) / \tau)}{\sum_{c \in \boldsymbol{C_{r}}}exp((-A^\text{cap}_{q, c}) / \tau)}
    \label{eq:cap_neg_weight}
\end{equation}


\noindent Finally, the query refinement is defined as follows:

\begin{equation}
    \boldsymbol{z}'_q = \alpha\boldsymbol{z}_q + \underbrace{\beta \boldsymbol{z}^\text{CLS}_q}_{\text{Positive vector}} - \underbrace{\gamma\sum_{j \in \boldsymbol{C_{r}}}\frac{w^\text{img}_{q, j}\boldsymbol{z}^\text{img}_j + w^\text{cap}_{q, j}\boldsymbol{z}^\text{cap}_j}{2}}_{\text{Negative vector}}
    \label{eq:rocchio_summ}
\end{equation}

\noindent Thus, weight $w^\text{img}_{q, j}$ defines the contribution of the image $j$ from the set of relevant items $\boldsymbol{C_r}$ to the negative relevance vector, based on its low cross-attention score with respect to query $q$. \revision{In Section \ref{sec:results_ablations}, we show the importance of using both positive and negative components.}

\section{Implementation Details}
\subsection{Datasets}
We use the Flickr30k \cite{young2014flickr} and COCO \cite{lin2014coco} datasets containing images and human-annotated captions. Each of these datasets contains 5 captions per image. Unless specified otherwise, we use one caption per image as a query, whereas the remaining captions are utilized to mimic explicit user feedback in one of the strategies further described in Section \ref{sec:strategies}. We utilize both datasets with the Karpathy splits widely used for retrieval tasks \cite{karpathy2015deep}.

\noindent\textbf{Flickr30k:} The dataset contains around 32000 images, divided into the train split of 30000 images and the validation and test splits with approximately 1000 images each.

\noindent\textbf{COCO:} The training data consists of more than 80000 images; the validation and test splits contain 5000 images.

The retrieval models are evaluated on the hold-out test splits of the datasets. In both datasets, each caption (user query) corresponds to only one image in the dataset. We evaluate our models using Hits@1, Hits@5, and MRR@5.\footnote{While Hits@1 and Hits@5 compute the fraction of queries where the correct image appears in the top-1 and top-5 results, MRR@5 captures the average rank position of the correct image within the top-5 results.}

\subsection{Synthetic Caption Generation}
We used LLaVA-1.5 \cite{liu2024improved}, quantized to 8-bit, to generate synthetic captions for images in both datasets. To obtain diverse yet concise image descriptions, we employed two-step prompting: the first prompt was randomly sampled from a pool of templates, whereas the second one constrained the model to adjust captions to be under 10 words. Details on caption generation are available in App.~\ref{sec:appendix_llava}.



\subsection{Retrieval Backbones}
\label{sec:vlm_models}

\revision{In this paper, we experiment with four pre-trained VLMs of different sizes. Namely, we use CLIP-ViT-B/32, CLIP-ViT-L/14 \cite{radford2021learning}, SigLIP (SoViT-400m) 
\cite{zhai2023sigmoid}, and BLIP-2 \cite{li2023blip} containing approximately 150M, 450M, 880M, and 1.2B parameters, respectively.} For all models, query-image relevance scores are computed as cosine similarity between the projected outputs of text and image encoders. Given that image features generated by BLIP-2 are two-dimensional\footnote{Q-Former in BLIP-2 fuses patch-level image representations into 32 query vectors.}, we use the largest similarity across the first dimension as the relevance score to the textual query.  

\subsection{Training Attentive Feedback Summarizer}
The AFS model was trained for 100 epochs on training sets with early stopping activated after 10 epochs without improvements in validation loss. \revision{We conducted all training and retrieval experiments on NVIDIA H100 GPUs.} We optimized model parameters using AdamW with an initial learning rate of 0.0003, weight decay of 0.01 and cosine annealing learning rate scheduling. \revision{We used batch sizes of 512, 256, 192, and 128 for CLIP-ViT-B/32, CLIP-ViT-L/14, SigLIP, and BLIP-2 (itm-vit-g). We set the number of attention heads in AFS to 12 for CLIP models and 16 for SigLIP and BLIP-2. We kept the dimensionality of the AFS equal to the output dimensionality of the retrieval backbones. These design choices affected the final AFS sizes per backbone VLM. 
Depending on the backbone, training on the Flickr30K dataset takes between 5 and 40 minutes per epoch, with convergence reached within 30–40 epochs. Table~\ref{tab:impl} reports the size of the AFS module with its inference latency for each backbone.}

\begin{table}[]
\centering
\scalebox{0.75}{
\begin{tabular}{@{}cccc@{}}
\toprule
Backbone & \# parameters (M) & \% of backbone & Latency (sec/item) \\ \midrule
CLIP-ViT-B/32   & 15.6 & 10.5\% & 0.03 \\
CLIP-ViT-L/14   & 16.1 & 3.6\% & 0.09 \\
SigLIP          & 25.2 &  2.9\% & 0.28 \\
BLIP-2          & 2.2  &  0.2\%  & 0.01 \\ \bottomrule
\end{tabular}}
\caption{\revision{\textbf{AFS sizes and latency per backbone.}}}
\label{tab:impl}
\end{table}

\subsection{Relevance Feedback Strategies}
\label{sec:strategies}

We conduct experiments with relevance feedback using two-turn and multi-turn retrieval. In the two-turn setting, the query is refined once using relevance feedback from items retrieved by the baseline model. In the multi-turn setting, the query is iteratively refined based on newly retrieved items. We define the retrieval baseline with VLMs and four strategies for relevance feedback.

\noindent\textbf{No feedback:} Zero-shot retrieval with VLMs and without relevance feedback. It serves as a baseline for comparison with other methods.

\noindent\textbf{Pseudo-relevance feedback (PRF):} Query representations are refined using Rocchio applied to the top-K retrieved images (Section~\ref{sec:meth_rocchio}). By default, we use $K=5$. Section \ref{sec:results_ablations} presents an ablation study for different values of K.

\noindent\textbf{Generative relevance feedback (GRF):} Query representations are refined using embeddings of synthetic captions corresponding to the top-K retrieved images (Section~\ref{sec:meth_grf}).

\noindent\textbf{Attentive feedback summarizer (AFS):} Relevance feedback is aggregated using the attentive feedback summarizer (Section~\ref{sec:meth_afs}). This model can be seen as a combination of PRF and GRF, simultaneously aggregating fine-grained information from the top-K retrieved images and their corresponding synthetic captions.

\noindent\textbf{Explicit feedback:} 
At each iteration of feedback aggregation, we average initial query embeddings with embeddings of a new query. The new user query is sampled from a pool of captions in a dataset corresponding to the same image as the initial query. Both datasets used in our study contain 5 captions per image, i.e., 5 user queries per image. In such a manner, we simulate explicit relevance feedback from a user who provides additional information. We use this strategy as a competitive baseline for other strategies that do not have access to ground truth.


\section{Results}

\subsection{Ablation Studies}
\label{sec:results_ablations}

We conduct ablations to evaluate different components and parameters using CLIP-ViT-B/32 for further analysis with all backbones. Specifically, we examine the impact of (1) the proposed Rocchio extension, (2) loss components in AFS, and (3) the importance of both positive and negative relevance feedback. \revision{In our experiments, we fix the number of items in relevance feedback $K = 5$ (App. \ref{sec:top-k}).}

\begin{figure}[!t]
    \centering
    \begin{minipage}{0.49\linewidth}
        \centering
        \includegraphics[width=\linewidth]{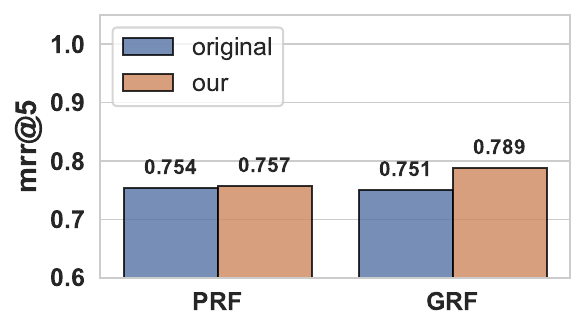}
        \\  (c) Flickr30K
    \end{minipage}
    \begin{minipage}{0.49\linewidth}
        \centering
        \includegraphics[width=\linewidth]{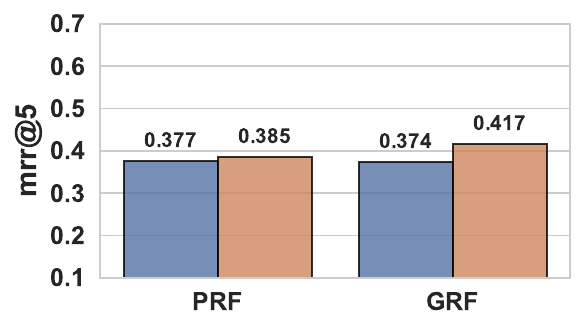}
        \\  (b) COCO
    \end{minipage}
    
    \caption{\textbf{Retrieval with Rocchio: original vs ours}. Retrieval metrics with relevance feedback using original and extended versions of Rocchio.}
    \label{fig:rocchio_ablation}
\end{figure}

\noindent\textbf{(1) Rocchio version:}
Further, we compare the proposed extension of the Rocchio algorithm (Section \ref{sec:meth_rocchio}) against its original counterpart (App. \ref{sec:appendix_rocchio}) with PRF and GRF. As shown in Figure \ref{fig:rocchio_ablation}, our extension of Rocchio positively affects the GRF strategy on both datasets. 
Regardless of the Rocchio version, the PRF strategy yields metrics comparable to the no feedback baseline, failing to effectively gather feedback from embeddings of the retrieved images. We will use our version of Rocchio for further experiments. Additionally, App. \ref{sec:appendix_temperature} evaluates the effect of temperature values $\tau$ (Equation 1) on retrieval performance.

\noindent\textbf{(2) Loss components in AFS:}
The proposed training objective for the AFS strategy (Equation~\ref{eq:summarizer_objective}) consists of two components: image loss and caption loss. The image loss enforces alignment with ground truth image embeddings, while the caption loss aligns summarized representations with additional ground truth captions. To isolate the contribution of each component, we train ablation variants with only one loss term enabled. Results for CLIP-ViT-B/32 with a single feedback aggregation turn are presented in Table~\ref{tab:ablation_losses}. Disabling image loss degrades performance across all evaluation metrics. In contrast, using only image loss (rows shaded in green) yields comparable Hits@5 and MRR@5 scores while outperforming the combined objective in Hits@1. This suggests that the summarizer can be trained effectively with ground-truth image embeddings only, reducing the need for multiple captions per image. For further evaluations, we will use the AFS variant trained with the image alignment loss component. \revision{One possible explanation is that aligning with image embeddings only reduces the multimodality gap \cite{liang2022mind}. While not providing direct evidence, Appendix~\ref{sec:appendix_afs_pca} presents PCA visualizations of the learned AFS representations, offering qualitative support for this hypothesis.}

\noindent\textbf{(3) Rocchio weights:}
According to Equations \ref{eq:rocchio_ours} and \ref{eq:rocchio_summ}, we fuse query, positive and negative vectors with weights $\alpha$, $\beta$ and $\gamma$, respectively. Here, we evaluate their effect on GRF and AFS. Table \ref{tab:ablation_feedback} shows that feedback aggregation with equal weights ($\alpha = \beta = \gamma$) results in poor performance for both feedback variants. Further, the performance is suboptimal when only one of the feedback signals ($\beta=0$ or $\gamma=0$) is enabled. Further in the paper, we use our methods with $\alpha=0.8$, $\beta=0.1$, $\gamma=0.1$, achieving decent performance for both GRF and AFS.

    


\definecolor{lightgreen}{RGB}{200, 255, 200}

\begin{table}[!t]
    \centering
    \scalebox{0.7}{ 
    \begin{tabular}{lcc|ccc}
        \toprule
        Dataset & $l^{\text{img}}_q$ & $l^{\text{cap}}_q$ & Hits@1 & Hits@5 & MRR@5 \\
        \midrule
        & \xmark & \cmark & 0.678 & 0.886 & 0.760 \\
        \rowcolor{lightgreen}
        Flickr30K & \cmark & \xmark & 0.724 & 0.913 & 0.801 \\
         & \cmark & \cmark & 0.708 & 0.910 & 0.791 \\
        \midrule
           & \xmark & \cmark & 0.296 & 0.531 & 0.383 \\
        \rowcolor{lightgreen}
        COCO   & \cmark & \xmark & 0.336 & 0.582 & 0.428 \\
           & \cmark & \cmark & 0.320 & 0.572 & 0.414 \\
        \bottomrule
    \end{tabular}
    }
    \caption{\textbf{Ablation on AFS loss components.} We evaluate AFS performance after training the model with different loss components. Image and caption loss components are denoted $l^{\text{img}}_q$ and $l^{\text{cap}}_q$ (Equations \ref{eq:img_loss}-\ref{eq:summarizer_objective}).}
    \label{tab:ablation_losses}
\end{table}

\begin{table}[!t]
    \centering
    \renewcommand{\arraystretch}{1.1}
    \setlength{\tabcolsep}{3pt}
    \scalebox{0.7}{  
    \begin{tabular}{lccc|ccc|ccc}
        \toprule
        \multirow{2}{*}{} & \multicolumn{3}{c|}{Rocchio weights} & \multicolumn{3}{c|}{Flickr} & \multicolumn{3}{c}{COCO} \\
        & $\alpha$ & $\beta$ & $\gamma$ & Hits@1 & Hits@5 & MRR@5 & Hits@1 & Hits@5 & MRR@5 \\
        \midrule
        \multirow{5}{*}{AFS} & 0.33 & 0.33 & 0.33 & 0.577 & 0.785 & 0.657 & 0.236 & 0.427 & 0.305 \\
        & 0.60 & 0.20 & 0.20 & \textbf{0.728} & {0.905} & {0.800} & \textbf{0.348} & \textbf{0.591} & \textbf{0.437} \\
        & 0.80 & 0.10 & 0.10 & {0.724}	& \textbf{0.913}	& \textbf{0.801}	& {0.336}	& {0.582}	& {0.428} \\
        & 0.80 & 0.00 & 0.20 & 0.594 & 0.773 & 0.662 & 0.215 & 0.411 & 0.285 \\
        & 0.80 & 0.20 & 0.00 & 0.706 & 0.906 & 0.786 & 0.320 & 0.559 & 0.410 \\
        \midrule
        \multirow{5}{*}{GRF} & 0.33 & 0.33 & 0.33 & 0.022 & 0.052 & 0.032 & 0.002 & 0.004 & 0.003 \\
        & 0.60 & 0.20 & 0.20 & 0.480 & 0.666 & 0.551 & 0.109 & 0.195 & 0.140 \\
        & 0.80 & 0.10 & 0.10 & \textbf{0.716} & \textbf{0.896} & \textbf{0.789} & \textbf{0.330} & \textbf{0.561} & \textbf{0.417} \\
        & 0.80 & 0.00 & 0.20 & 0.455 & 0.631 & 0.522 & 0.098 & 0.171 & 0.125 \\
        & 0.80 & 0.20 & 0.00 & 0.637 & 0.890 & 0.738 & 0.279 & 0.524 & 0.370 \\
        \bottomrule
    \end{tabular}
    }
    \caption{\textbf{Ablation on positive and negative feedback.} Performance metrics for different weights for query, positive, and negative vectors.}
    \label{tab:ablation_feedback}
\end{table}

\begin{table*}[!t]
    \centering
    \renewcommand{\arraystretch}{1}
    \setlength{\tabcolsep}{7pt}
    \scalebox{0.7}{ 
    \begin{tabular}{lll|ccc|ccc}
        \toprule
        \multicolumn{3}{c|}{} & \multicolumn{3}{c|}{\textbf{Flickr30k}} & \multicolumn{3}{c}{\textbf{COCO}} \\
        \midrule
        Row & \textbf{Retrieval backbone} & \textbf{Feedback} & \textbf{Hits@1} & \textbf{Hits@5} & \textbf{MRR@5} & \textbf{Hits@1} & \textbf{Hits@5} & \textbf{MRR@5} \\
        \midrule
        1  &  & No feedback & 0.671 & 0.890 & 0.758 & 0.295 & 0.542 & 0.385 \\
        2  &  & PRF         & 0.669 & 0.892 & 0.757 & 0.295 & 0.541 & 0.385 \\
        3  & CLIP-ViT-B/32 & GRF         & 0.716 & 0.896 & 0.789 & \underline{0.330} & 0.561 & \underline{0.417} \\
        4  &  & AFS         & \underline{0.724}	& \underline{0.913}	& \underline{0.801}	& \textbf{0.336}	& \textbf{0.582}	& \textbf{0.428} \\
        5  &  & Explicit    & \textbf{0.725} & \textbf{0.937} & \textbf{0.809} & 0.316 & \underline{0.574} & 0.411 \\
        \midrule
        6  &  & No feedback & 0.727 & 0.912 & 0.800 & 0.347 & 0.594 & 0.439 \\
        7  &  & PRF         & 0.726 & 0.917 & 0.800 & 0.347 & 0.593 & 0.437 \\
        8  & CLIP-ViT-L/14 & GRF         & \underline{0.766} & 0.928 & \underline{0.833} & 0.373 & 0.596 & 0.457 \\
        9  &  & AFS         & \textbf{0.784} & \textbf{0.943} & \textbf{0.846} & \textbf{0.386} & \underline{0.621} & \textbf{0.475} \\
        10 &  & Explicit    & 0.746 & \underline{0.933} & 0.820 & \underline{0.378} & \textbf{0.623} & \underline{0.470} \\
        \midrule
        11 &    & No feedback & 0.868 & 0.975 & 0.915 & 0.543 & 0.794 & 0.639 \\
        12 &    & PRF         & 0.869 & 0.976 & 0.917 & 0.541 & 0.791 & 0.636 \\
        13 & BLIP-2   & GRF         & \textbf{0.893} & \textbf{0.985} & \textbf{0.932} & \underline{0.562} & 0.792 & 0.649 \\
        14 &   & AFS         & \underline{0.890} & 0.982 & \underline{0.931} & 0.558 & \underline{0.797} & \underline{0.651} \\
        15 &    & Explicit    & 0.886 & \underline{0.984} & 0.928 & \textbf{0.583} & \textbf{0.825} & \textbf{0.677} \\
        \bottomrule
    \end{tabular}
    }
    \caption{\textbf{Two-turn retrieval.} Retrieval performance with five relevance feedback benchmarks after one round of feedback aggregation. The two highest metric values per dataset and backbone model are highlighted in \textbf{bold} and \underline{underlined}.}
    \label{tab:benchmark}
\end{table*}

\subsection{Evaluating Relevance Feedback Strategies}
\label{sec:benchmarking_res}

We evaluate the proposed relevance feedback strategies defined in Section \ref{sec:strategies} with the VLM backbones (Section \ref{sec:vlm_models}) using the configurations identified in ablations.

\noindent\textbf{Two-turn retrieval with relevance feedback:} In Table \ref{tab:benchmark}, we present the metrics obtained in two-step retrieval, i.e., when the query representations were refined once after initial retrieval. First, AFS, GRF, and explicit feedback consistently outperform the no feedback baseline. Notably, CLIP-ViT-B/32 with AFS (row 4) nearly matches the baseline performance of CLIP-ViT-L/14 (row 6), which is three times larger. Comparing AFS to GRF for CLIP models, we note that the former generally shows higher Hits@5. Given that we used top-5 ranked items as relevance feedback for both AFS and GRF,
this suggests that AFS more effectively aggregates feedback from these items, even when the matching image was not one of them. For the larger BLIP-2 model, AFS, GRF, and explicit feedback perform comparably on Flickr30K, while explicit feedback yields larger gains on COCO. Finally, PRF does not show noteworthy improvements and yields results comparable to the no feedback baseline.


\begin{table}[]
\centering
\scalebox{0.7}{
\begin{tabular}{lcccc}
\hline
\multicolumn{1}{c}{}                   & \multicolumn{2}{c}{Flickr30K}                & \multicolumn{2}{c}{COCO}                     \\
\multicolumn{1}{c}{}                   & Hits@5               & $\Delta$ & Hits@5               & $\Delta$ \\ \hline
FLAME \cite{cao2025flame}         & 91.7                 & -                     & 70.4                 & -                     \\
FLAIR \cite{xiao2025flair}         & 94.9                 & -                     & 77.5                 &    -                   \\ \hline
SigLIP + VISA \cite{guofengding2025visual} & 97.1                 & +1.0                  & 80.3                 & +3.5                  \\
BLIP-2 + VISA \cite{guofengding2025visual} & 98.4                 & +0.3                  & 88.0*                & +0.3                  \\ \hline
CLIP-L + AFS (ours)                    & 94.3                 & +3.1                  & 62.1                 & +2.8                  \\
SigLIP + AFS (ours)                    & 96.8 &  +0.6 & 77.9 & +0.3  \\
BLIP-2 + AFS (ours)                    & 98.2                 & +0.7                  & 79.7                 & +0.3                 
\end{tabular}
}
\caption{\revision{\textbf{Comparing AFS with recent approaches.} Values are reported in percentage points. For test-time methods, we present the scores after applying them, along with the corresponding performance change $\Delta$. * -- BLIP-2 was fine-tuned on the COCO dataset before applying VISA \cite{guofengding2025visual}.}}
\label{tab:sota}
\end{table}

\revision{Moreover, Table \ref{tab:sota} compares the performance of AFS against recent methods from the literature, namely FLAME \cite{cao2025flame}, FLAIR \cite{xiao2025flair}, and VISA \cite{guofengding2025visual}. While FLAIR and FLAME improve the retrieval backbone through model training, VISA, similarly to AFS, operates as an additional module at inference time. As can be seen, zero-shot BLIP-2 with AFS and BLIP-2 with VISA show the highest Hits@5 scores on the Flickr30K dataset, whereas VISA applied on top of the fine-tuned BLIP-2 achieves the highest performance on COCO. Note that in our work, we applied AFS to backbones in a zero-shot manner, i.e., we did not fine-tune the VLMs. In most cases, except for SigLIP on COCO, AFS and VISA demonstrate comparable performance gains ($\Delta$) when applied on top of different retrieval backbones across both datasets. In terms of computational overhead at inference time, VISA relies on pre-trained VLMs and LLMs, which cumulatively contain billions of parameters for the most performing VISA variant reported in Table \ref{tab:sota}. In turn, while requiring training, the AFS modules are significantly more compact, containing between 2M and 25M parameters depending on the VLM backbone.} 

Additionally, an experiment reported in App. \ref{sec:appendix_afs-prf} demonstrates that AFS operating with relevant images only (without synthetic captions) shows comparable performance to its full variant.

\begin{figure}[!t]
    \centering
    \begin{minipage}{0.32\linewidth}
        \centering
        \includegraphics[width=\linewidth]{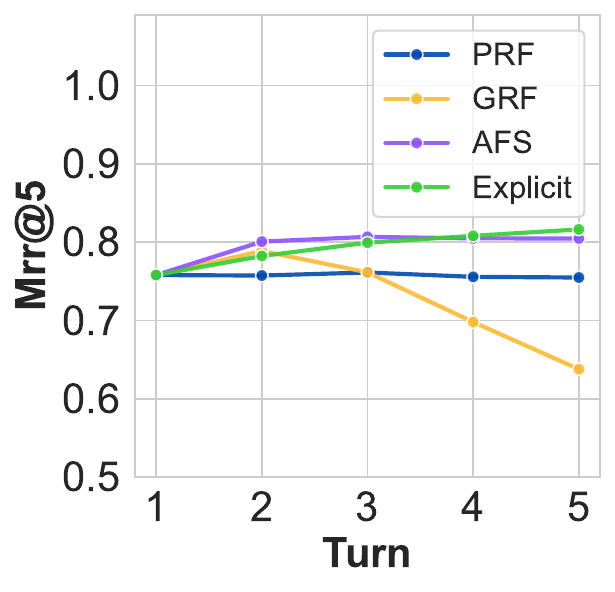}
        \\ (a) CLIP-B
        \label{fig:multiturn_clip_base}
    \end{minipage}
    \begin{minipage}{0.32\linewidth}
        \centering
        \includegraphics[width=\linewidth]{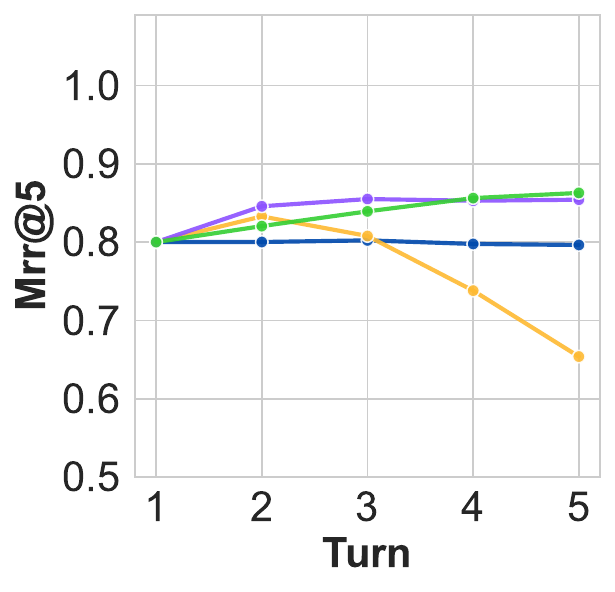}
        \\ (b) CLIP-L
        \label{fig:multiturn_clip_large}
    \end{minipage}
    \begin{minipage}{0.32\linewidth}
        \centering
        \includegraphics[width=\linewidth]{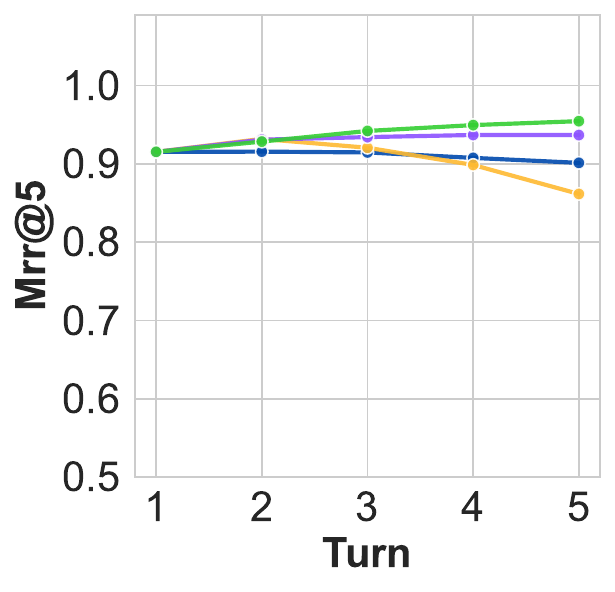}
        \\ (c) BLIP-2
        \label{fig:fig:multiturn_blip2}
    \end{minipage}
    
    \caption{\textbf{Multi-turn retrieval on Flickr30K.} MRR@5 scores for multi-turn retrieval with relevance feedback. CLIP-B and CLIP-L refer to CLIP-ViT-B/32 and CLIP-ViT-L/14, respectively.}
    \label{fig:multiturn_flickr}
\end{figure}

\noindent\textbf{Multi-turn feedback aggregation:} Relevance feedback supports multi-turn retrieval through iterative query refinement. While real-world constraints may limit the number of turns, we aim to study whether iterative refinement consistently improves results or if multiple rounds of feedback provide little additional benefit. For PRF, GRF, and AFS, we use embeddings of the top-5 items retrieved in the preceding round as relevance feedback. For explicit feedback, at each turn, we fuse representations of another caption from ground truth annotations.

Figure~\ref{fig:multiturn_flickr} shows MRR@5 scores obtained in such settings on Flickr30K. The explicit feedback strategy shows consistent performance growth, which is expected due to the refinement with ground-truth user captions. In contrast, the performance of GRF begins to decline starting from the third round of retrieval, which may be linked to the query drift phenomenon discussed in the relevance feedback literature \cite{ruthven2003survey, zighelnic2008query}. 
Finally, AFS shows a positive performance trend across retrieval rounds, avoiding the drift, although it underperforms compared to explicit feedback in the later rounds. The full version of the multi-turn retrieval results for both datasets is presented in App. \ref{sec:appendix_multiturn}.

    

\subsection{Visualizing Feedback Summarization}
\label{sec:results_explore_afs}

\begin{figure}[t]
    \centering
    \includegraphics[width=0.8\linewidth]{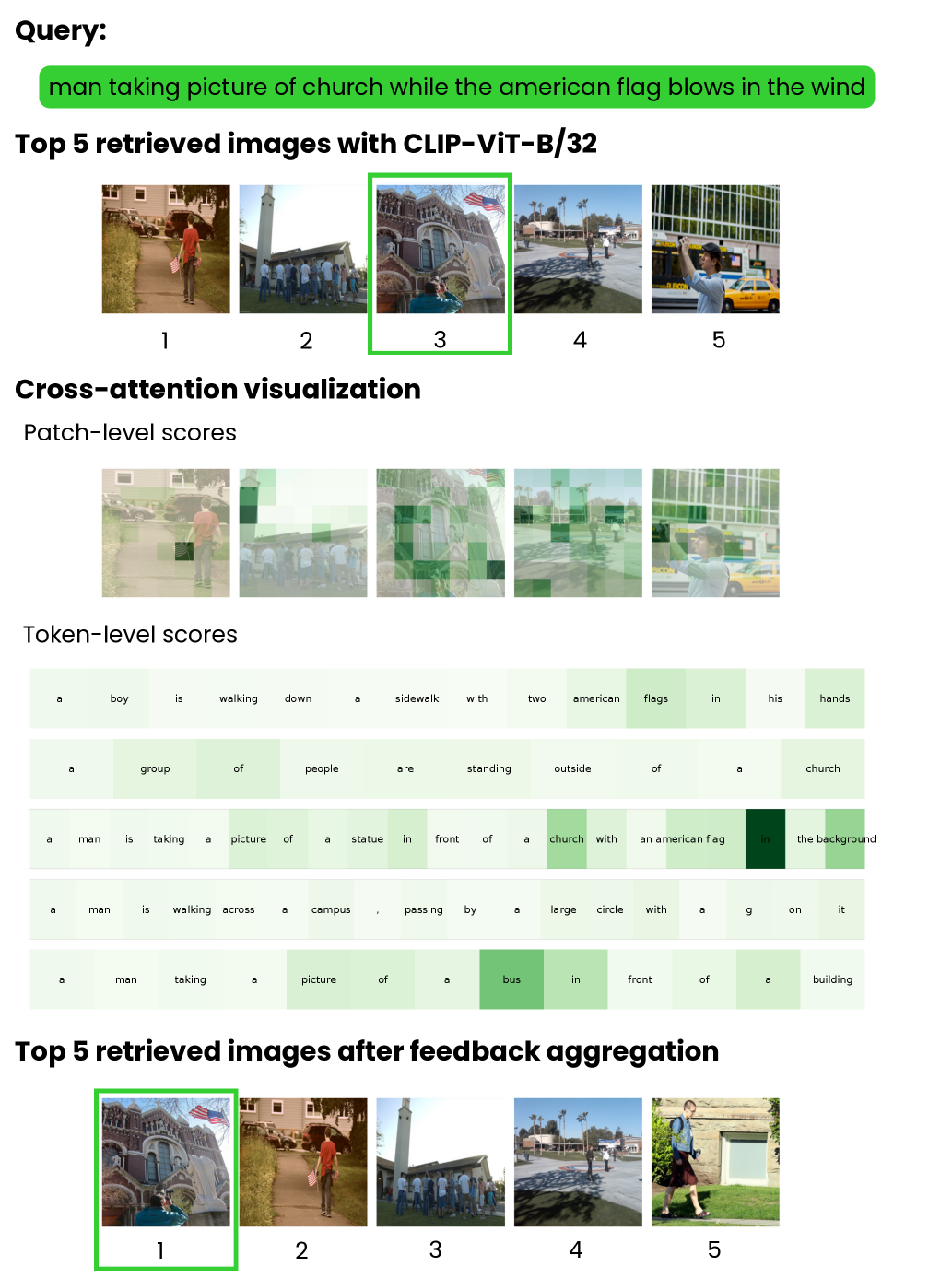}
    \caption{\textbf{Cross-attention visualization.} The example is sampled from Flickr30K dataset and processed with CLIP-ViT-B/32 as a retrieval backbone. The ground truth image corresponding to the query is highlighted with a green frame.}
    \label{fig:summarizer_attention}
\end{figure}

Beyond improving retrieval performance, 
a unique advantage of the AFS architecture (Figure~\ref{fig:summarizer_architecture}) over other feedback strategies is its ability to generate saliency maps that visualize how feedback is aggregated. By analyzing cross-attention scores, we can assess the contribution of each patch in retrieved images and each token in synthetic captions toward query refinement---a level of insight not available with PRF or GRF.\footnote{Note: PRF and GRF include only global image embeddings, so we can only generate global relevance scores (i.e., cosine similarities) for each item. Therefore, PRF and GRF can produce the information in Figure \ref{fig:summarizer_attention}, but without patch- and word-level saliency maps.} Specifically, we aggregate the cross-attention weights for each item in the relevance sequence and normalize them between 0 and 1. These scores allow us to visualize which parts of relevant images and captions the summarizer prioritizes during refinement (see App.~\ref{sec:appendix_afs_aggregation} for details). We illustrate an example of cross-attention saliency maps in Figure~\ref{fig:summarizer_attention}. Aggregated attention scores highlight semantically meaningful regions related to the query. 
While some highlighted areas align with user intent, we also observe cases where attention is assigned to less intuitive regions. Notably, the aggregation of feedback with AFS results in improved retrieval performance, with the correct matching image ranked first. 
Additional examples of attention visualization are provided in App.~\ref{sec:appendix_vis}.

\section{Discussion}
\label{sec:discussion}
In this section, we discuss the impact of relevance feedback on retrieval performance and its implications for interactive retrieval. Further, limitations are discussed in App.~\ref{sec:appendix_limitations}.

\noindent\textbf{Impact of relevance feedback:} Our evaluations show that applying classical pseudo-relevance feedback alone, such as the standard Rocchio method, is not sufficient. We introduced more effective strategies to enable relevance feedback for image search---GRF with an extended Rocchio formulation, and AFS---and carefully evaluated different design choices through a series of ablations (Section~\ref{sec:results_ablations}).

In Section~\ref{sec:benchmarking_res}, we demonstrate that these methods can improve retrieval performance and substantially narrow the performance gap between smaller and larger VLMs. For example, CLIP-ViT-B/32 combined with AFS nearly matches the baseline performance of CLIP-ViT-L/14, even though the latter is approximately three times larger (Table~\ref{tab:benchmark}). Thus, relevance feedback provides a practical alternative to scaling up VLM size and improves retrieval performance with larger models, such as BLIP-2.

\noindent\textbf{Trade-offs:} Among the introduced techniques, explicit feedback often achieves the highest performance, but it relies on textual inputs from users. When such interaction is not feasible, GRF and AFS serve as effective alternatives. GRF can be effective in two-turn retrieval but requires synthetic image captions for every candidate image, adding processing time in the background. In contrast, AFS can function without synthetic captions (App.~\ref{sec:appendix_afs-prf}) and show competitive retrieval performance. However, despite its compact design, AFS incurs additional inference overhead for forward propagation. These trade-offs highlight the importance of considering application-specific constraints when integrating relevance feedback.

\noindent\textbf{Towards interactive retrieval:} While our methods were evaluated in an offline setting, both GRF and AFS offer mechanisms that are well-suited for integration into interactive retrieval interfaces. GRF refines query embeddings based on textual relevance signals (e.g., synthetic captions), and AFS summarizes visual feedback at a fine-grained level using attention mechanisms. Despite varying in the form of feedback, both methods are integrated into a shared Rocchio-style embedding update. \revision{Besides, we used simulated explicit feedback as an upper baseline for AFS and GRF. While this approach sets a strong performance benchmark, it does not fully represent real user interactions. Future work should therefore explore how user interactions with retrieved items (e.g., clicks, region selection, dwell time)
can be integrated with the proposed relevance feedback strategies for user-guided interactive retrieval.} In App. \ref{sec:appendix_demo}, we present an initial demonstration showing how user feedback on retrieved items can be incorporated in AFS.


\section{Conclusion}
We designed and systematically evaluated relevance feedback strategies for text-to-image retrieval using modern vision-language models. We compared the proposed strategies to each other and zero-shot VLM-based retrieval. Building on the Rocchio algorithm, we explored pseudo-relevance feedback (PRF) and generative relevance feedback (GRF) using retrieved images and AI-generated captions. Additionally, we introduced an attentive feedback summarizer (AFS) that aggregates fine-grained signals from images and captions. This compact transformer-based model outperforms both PRF and GRF, showing robust performance in multi-turn scenarios, though it still lags behind explicit feedback.
Our findings show that relevance feedback is a promising technique to improve retrieval performance across a range of vision-language models, narrowing the gap between the smaller and larger VLM variants. 
 




\clearpage



{
    \small
    \bibliographystyle{ieeenat_fullname}
    \bibliography{main}
}

\clearpage

\section*{Appendix}
\appendix

\section{Original Rocchio formulation}
\label{sec:appendix_rocchio}
The Rocchio algorithm \cite{rocchio1971relevance} aims to generate the refined query vector $\boldsymbol{z}'_q \in \mathbb{R}^d$, by fusing representations of original query $\boldsymbol{z}_q$ with relevant and irrelevant (or positive and negative) feature vectors $\boldsymbol{z}_p$ and $\boldsymbol{z}_n$ obtained in the current stage of retrieval. This updated representation is then used to perform another round of retrieval. The Rocchio query refinement rule can formally be defined as follows:

\begin{equation}
    \boldsymbol{z}'_q = \alpha  \boldsymbol{z}_q + \beta  \boldsymbol{z}_p - \gamma \boldsymbol{z}_n
    \label{eq:rocchio_main}
\end{equation}

The positive vector is computed from the representations of the most relevant candidates, whereas the negative features are aggregated from the irrelevant ones. Specifically, the aggregated positive and negative representations are computed by averaging:

\begin{equation}
    \boldsymbol{z}_p = \frac{1}{|\boldsymbol{C_r}|}\sum_{i \in \boldsymbol{C_r}}\boldsymbol{z}_i
\end{equation}
\begin{equation}
    \boldsymbol{z}_n = \frac{1}{|\boldsymbol{C_{nr}}|}\sum_{i \in \boldsymbol{C_{nr}}}\boldsymbol{z}_i
\end{equation}

\noindent where $\boldsymbol{C_r}$ and $\boldsymbol{C_{nr}}$ are the sets of the most relevant and non-relevant candidates with cardinalities $|\boldsymbol{C_r}|$ and $|\boldsymbol{C_{nr}}|$. In practice, the algorithm is often applied with $K = |\boldsymbol{C_r}| = |\boldsymbol{C_{nr}}|$, i.e., using the top-K candidates with the highest and lowest relevance with respect to the query. 


\section{Generating Synthetic Captions with LLaVA-1.5}
\label{sec:appendix_llava}

\begin{figure}[!ht]
    \centering
    \begin{minipage}{0.8\linewidth}
        \centering
        \includegraphics[width=\linewidth]{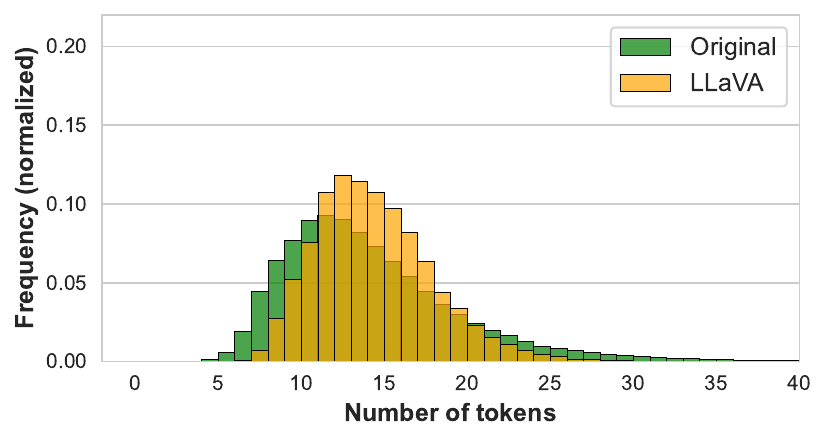}
        \\ (a) Flickr30K
        \label{fig:llava_flickr_length}
    \end{minipage}
    \begin{minipage}{0.8\linewidth}
        \centering
        \includegraphics[width=\linewidth]{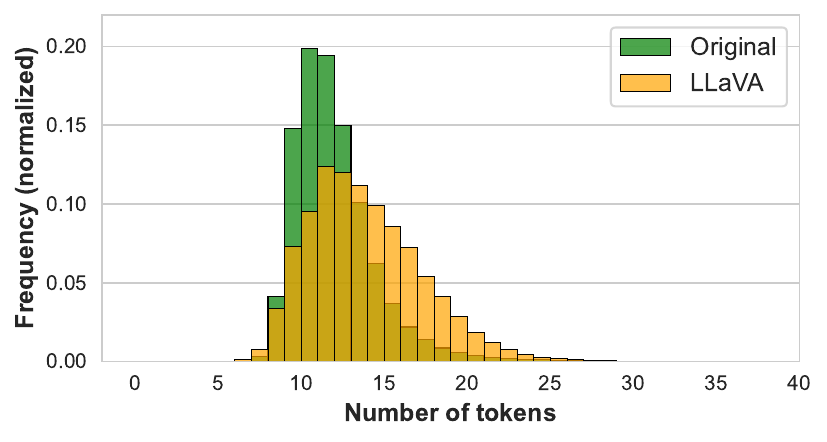}
        \\ (b) COCO
        \label{fig:llava_coco_length}
    \end{minipage}
    
    \caption{Distribution of original and LLaVA-generated caption lengths.}
    \label{fig:llava_length}
\end{figure}

We utilized LLaVA-1.5-7B\footnote{We used open-source checkpoints available via HuggingFace: https://huggingface.co/llava-hf/llava-1.5-7b-hf} \cite{liu2024improved}, quantized in 8-bit, to generate a synthetic caption for each image in both Flickr30K and COCO. Given that these captions would later be used to obtain CLIP and BLIP-2 representations -- which typically rely on short and descriptive textual inputs \cite{rodriguez2024does, zhang2024long} -- we adopted a two-step prompting strategy to balance diversity and conciseness. First, we randomly selected a prompt from a predefined pool of five sentence-level description prompts, ensuring variability in the generated responses. Second, we provided a follow-up prompt explicitly instructing the model to generate a single-sentence description. The prompts and conversation template are shown in Table \ref{tab:llava_prompting}. Figure \ref{fig:llava_length} shows the distributions of the lengths of the original and generated captions. As observed, the length distribution of generated captions differs slightly from that of human-written ones while still avoiding extreme values and remaining within a reasonable range.

\begin{figure}[!t]
    \centering
    \begin{minipage}{0.49\linewidth}
        \centering
        \includegraphics[width=\linewidth]{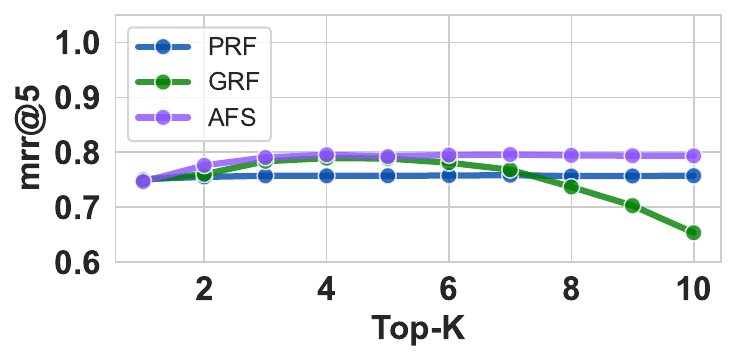}
        \\  (a) Flickr30K
    \end{minipage}
    \begin{minipage}{0.49\linewidth}
        \centering
        \includegraphics[width=\linewidth]{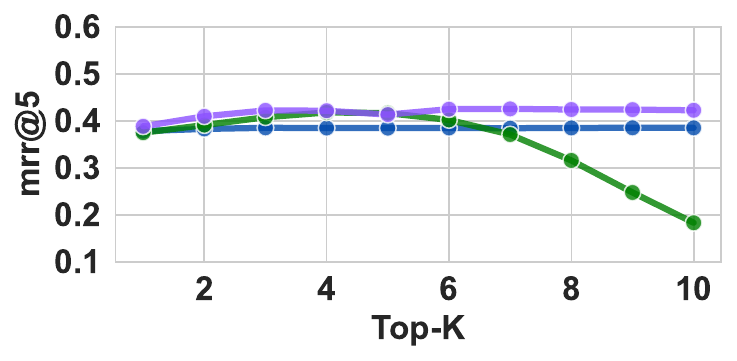}
        \\  (b) COCO
    \end{minipage}
    
    \caption{\textbf{Retrieval with Rocchio: top-K items}. Retrieval metrics with a varying number of items in relevance feedback.}
    \label{fig:topk_ablation}
\end{figure}

\begin{table}[!h]
\centering
\scalebox{0.8}{
\begin{tabular}{llcccc}
\toprule
\textbf{Dataset} & \textbf{Feedback} & \textbf{0.05} & \textbf{0.10} & \textbf{0.25} & \textbf{0.50} \\
\midrule
Flickr & GRF & 0.716 & 0.714 & 0.707 & 0.709 \\
          & PRF             & 0.669 & 0.670 & 0.669 & 0.671 \\
COCO      & GRF & 0.330 & 0.331 & 0.327 & 0.327 \\
          & PRF          & 0.295 & 0.295 & 0.295 & 0.294 \\
\bottomrule
\end{tabular}}
\caption{Retrieval performance (Hits@1) with different values of temperature.}
\label{tab:temperature_feedback}
\end{table}

\begin{table*}[!t]
    \centering
    \renewcommand{\arraystretch}{1.2}
    \scalebox{0.9}{
    \begin{tabular}{|c|l|p{10cm}|}
        \hline
        \textbf{Step} & \textbf{Role} & \textbf{Prompt} \\  
        \hline
        0 & \textbf{} & \texttt{<prompt1>} is randomly selected from the pool:  
        \begin{itemize}
            \item "Describe the image and main visual features in one sentence."
            \item "Generate a short caption for this image, focusing on the visual details."
            \item "Provide a concise description, one sentence, of the image visual features and surroundings."
            \item "Write a brief caption, one sentence, that describes the image visual features."
            \item "Summarize the image visual features in a precise caption, maximum one sentence."
        \end{itemize} 
        \texttt{<prompt2>}:  
        \begin{itemize}
            \item "Pay attention to the visual settings and details on the image. Write exactly one sentence under 10 words."
        \end{itemize} \\
        \hline \hline
        1 & \textbf{User} & \texttt{USER: <image><prompt1> ASSISTANT:} \\
 
        & \textbf{Assistant} & \texttt{USER: <image><prompt1> ASSISTANT: <answer1>} \\
        \hline \hline
        2 & \textbf{User} & \texttt{USER: <image><prompt1> ASSISTANT: <answer1>USER: <prompt2> ASSISTANT:} \\  

        & \textbf{Assistant} & \texttt{USER: <image><prompt1> ASSISTANT: <answer1>USER: <prompt2> ASSISTANT: <answer2>} \\
        \hline
    \end{tabular}}
    \caption{Conversation template for two-step prompting: \texttt{<answer2>} is the generated synthetic caption for input image \texttt{<image>}.}
    \label{tab:llava_prompting}
\end{table*}

\section{Rocchio Temperature}
\label{sec:appendix_temperature}

In Equation \ref{eq:weights}, we define positive and negative weights for each item in relevance feedback. Specifically, we scale similarities between each item and the corresponding query representation by temperature $\tau$ to compute the weight of each item. These weights are then used in Equation \ref{eq:rocchio_ours} to guide query representations. Table \ref{tab:temperature_feedback} shows that using lower temperature values, which sharpen the weight distribution, leads to better retrieval performance.

\section{Number of Candidates for Relevance Feedback}
\label{sec:top-k}
We evaluate the impact of the number of items in relevance feedback. According to Figure \ref{fig:topk_ablation}, the optimal performance in MRR@5 for both GRF and AFS is achieved with 4-5 relevant items on both datasets. Further increase of the relevance set leads to decay in MRR for GRF.

\section{Attentive Feedback Summarizer}
\label{sec:appendix_afs}

\subsection{Attention Blocks Architecture}
\label{sec:appendix_afs_att}

Section \ref{sec:meth_afs} presents an overview of the proposed attentive feedback summarizer architecture. This section provides a detailed description of how image patches and caption tokens are processed by the summarizer.

\begin{figure*}[h]
    \centering
    \includegraphics[width=0.7\linewidth]{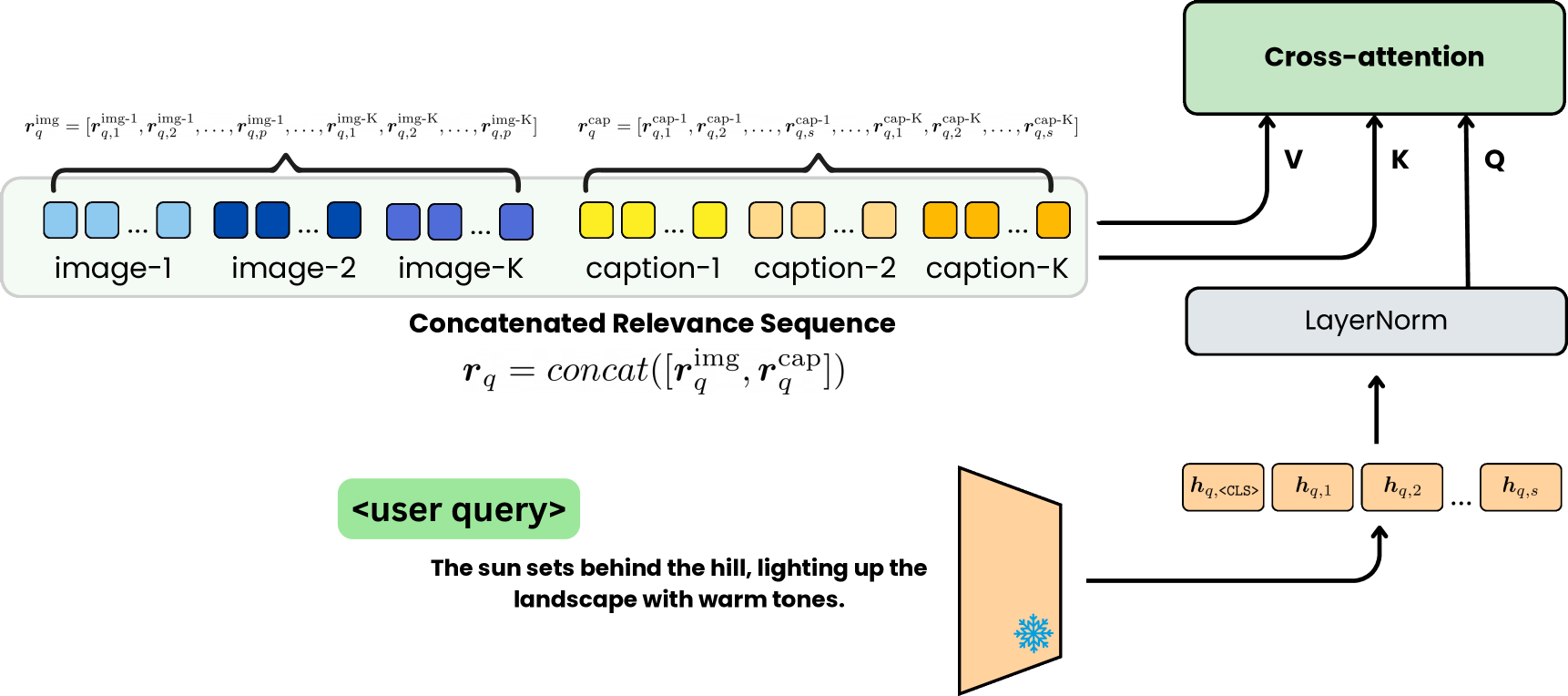}
    \caption{\textbf{Cross-attention architecture.}}
    \label{fig:cross-attention}
\end{figure*}

The first component of the AFS architecture is a cross-attention module (Figure \ref{fig:cross-attention}) designed to fuse query representations with relevance feedback. Specifically, cross-attention queries are obtained for the token features of the user query. We denote these token-level representations of input query $q$ with $s$ tokens as $\boldsymbol{h}_q = [\boldsymbol{h}_{q, \texttt{CLS}}, \boldsymbol{h}_{q,1}, \boldsymbol{h}_{q, 2}, \dots, \boldsymbol{h}_{q, j}, \dots, \boldsymbol{h}_{q, s}] \in \mathbb{R}^{s \times d_t}$, where $d_t$ is the token dimensionality and $\boldsymbol{h}_{q, \texttt{CLS}}$ is a learnable \texttt{CLS} token. Furthermore, keys and values are obtained from relevance feedback, comprising patch- and word-level representations of the most relevant images and their corresponding AI-generated captions. Relevance feedback vector $\boldsymbol{r}_q = concat([\boldsymbol{r}^{\text{img}}_{q}, \boldsymbol{r}^{\text{cap}}_{q}])$ is a concatenated sequence of the patch and token embeddings for the top-K relevant images and their synthetic captions projected to $\mathbb{R}^{d_t}$. In detail, visual relevance sequence can be defined as 
 $\boldsymbol{r}^{\text{img}}_{q} = [\boldsymbol{r}^{\text{img-1}}_{q, 1}, \boldsymbol{r}^{\text{img-1}}_{q, 2}, \dots, \boldsymbol{r}^{\text{img-1}}_{q, p}, \dots, \boldsymbol{r}^{\text{img-K}}_{q, 1}, \boldsymbol{r}^{\text{img-K}}_{q, 2}, \dots, \boldsymbol{r}^{\text{img-K}}_{q, p}]$
 for images divided into $p$ patches $\boldsymbol{r}^{\text{img-c}}_{q,j} \in \mathbb{R}^{d_t}$. Similarly, the word-level relevance sequence from corresponding AI-generated captions is $\boldsymbol{r}^{\text{cap}}_{q} = [\boldsymbol{r}^{\text{cap-1}}_{q, 1}, \boldsymbol{r}^{\text{cap-1}}_{q, 2}, \dots, \boldsymbol{r}^{\text{cap-1}}_{q, s}, \dots, \boldsymbol{r}^{\text{cap-K}}_{q, 1}, \boldsymbol{r}^{\text{cap-K}}_{q, 2}, \dots, \boldsymbol{r}^{\text{cap-K}}_{q, s}]$ with captions padded to length $s$ and $\boldsymbol{r}^{\text{cap-c}}_{q,j} \in \mathbb{R}^{d_t}$.
 
 The second component of the summarizer is a standard self-attention block that communicates information between query sequence tokens. Therefore, the output sequence of these two blocks can be defined as $\boldsymbol{\tilde{h}}_q = SelfAttn(CrossAttn(\boldsymbol{h}_q, \boldsymbol{r}_q))\in \mathbb{R}^{s\times {d_t}}$. We use \texttt{CLS} token $\boldsymbol{\tilde{h}}_{q,\texttt{CLS}}$ as the output of the feedback summarizer and apply linear projection to obtain $\boldsymbol{z}^{\texttt{CLS}}_q \in \mathbb{R}^d$.

\subsection{AFS Representations}
\label{sec:appendix_afs_pca}

\begin{figure}[!t]
    \centering
    \begin{minipage}{0.4\linewidth}
        \centering
        \includegraphics[width=\linewidth]{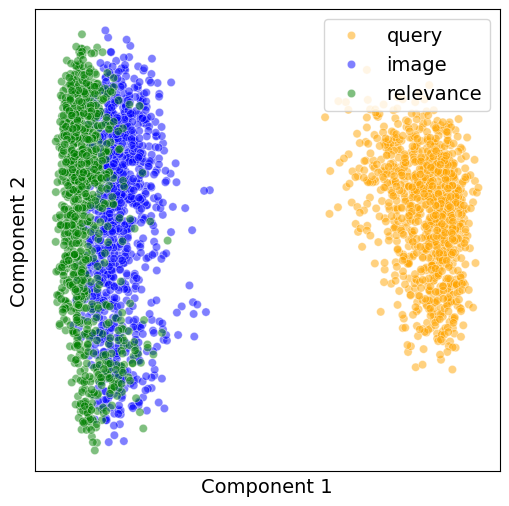}
        \\ (a) Image loss
        \label{fig:pca_img}
    \end{minipage}
    \begin{minipage}{0.4\linewidth}
        \centering
        \includegraphics[width=\linewidth]{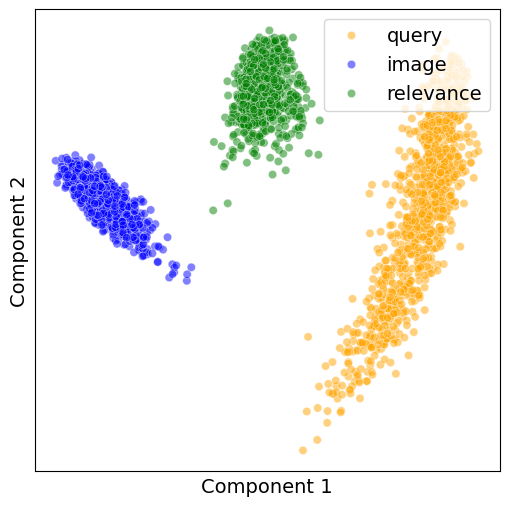}
        \\ (b) Combined loss
        \label{fig:pca_combined}
    \end{minipage}
    
    \caption{Images, queries and AFS representations projected with PCA.}
    \label{fig:pca}
\end{figure}

In this section, we visualize representations $\boldsymbol{z}^{\texttt{CLS}}_q$ learned by the AFS model. We compare two AFS variants trained with different objectives: one using only the image-based loss ($l_q^{img}$), and one using a combined loss from both image and caption supervision ($l_q^{img} + l_q^{cap}$), corresponding to rows 2 and 3 in Table~\ref{tab:ablation_losses}. As shown in Section~\ref{sec:results_ablations}, the variant trained with image-based loss achieves better retrieval performance. Figure~\ref{fig:pca} shows PCA projections of the query, image and AFS embeddings in two-dimensional space for Flickr30K with CLIP-ViT-B/32. The embeddings from the combined-loss model are positioned between the text and image embedding clusters, while the image-only variant produces representations that remain closer to the image features. This pattern suggests that feedback representations more aligned with image features may be more effective for refining query representations.

\subsection{Cross-attention Score Aggregation}
\label{sec:appendix_afs_aggregation}

In Section \ref{sec:results_explore_afs} and Appendix \ref{sec:appendix_vis}, we visualize cross-attention scores to demonstrate feedback summarization from fine-grained feedback. 

In this section, we provide technical details on cross-attention score aggregation. Cross-attention scores are defined as $\boldsymbol{A}_q \in \mathbb{R}^{n_h \times s_q \times s_r}$, where $n_h$ is the number of attention heads, $s_q$ is the number of tokens in a user query, and $s_r$ is the number of patch and token embeddings in a relevance sequence. First, we accumulate attention scores over attention heads and query tokens obtaining $\boldsymbol{\hat{A}}_q = [\hat{A}_{q, 1}, \hat{A}_{q, 2}, \dots, \hat{A}_{q, {s_r}}]\in \mathbb{R}^{s_r}$, where each item is a scalar corresponding to a certain patch or token in relevance sequence $\boldsymbol{r}_q$. Then, we split $\boldsymbol{\hat{A}}_q$ into image and caption parts 
$\boldsymbol{\hat{A}}^{\text{img}}_{q} = [\hat{A}^{\text{img-1}}_{q, 1}, \dots, \hat{A}^{\text{img-1}}_{q, p}, \dots, \hat{A}^{\text{img-K}}_{q, 1}, \dots, \hat{A}^{\text{img-K}}_{q, p}]$ 
and 
$\boldsymbol{\hat{A}}^{\text{cap}}_{q} = [\hat{A}^{\text{cap-1}}_{q, 1}, \dots, \hat{A}^{\text{cap-1}}_{q, s}, \dots, \hat{A}^{\text{cap-K}}_{q, 1}, \dots, \hat{A}^{\text{cap-K}}_{q, s}]$ for $K$ items used as relevance feedback. Finally, both sequences $\boldsymbol{\hat{A}}^{\text{img}}_{q}$ and $\boldsymbol{\hat{A}}^{\text{cap}}_{q}$ are independently normalized between 0 and 1 using min-max scaling. As a result, the normalized values are used as saliency scores for patches on relevant images and words in synthetic captions.

\subsection{AFS without Synthetic Captions}
\label{sec:appendix_afs-prf}

As shown in Table~\ref{tab:benchmark}, PRF does not improve over baseline retrieval models without relevance feedback (Table~\ref{tab:benchmark}: rows 2, 7, and 12). In Sections~\ref{sec:meth_afs} and~\ref{sec:strategies}, we introduced AFS as a strategy that combines PRF and GRF by aggregating local information from both image patches and synthetic captions. Here, we evaluate whether AFS can operate using retrieved image embeddings without synthetic captions. We refer to this variant as AFS-PRF.\footnote{We did not re-train the models using images only; instead, we used the same trained models from previous sections while ignoring synthetic captions at inference.}

\begin{figure}[!ht]
    \centering
    \begin{minipage}{0.49\linewidth}
        \centering
        \includegraphics[width=\linewidth]{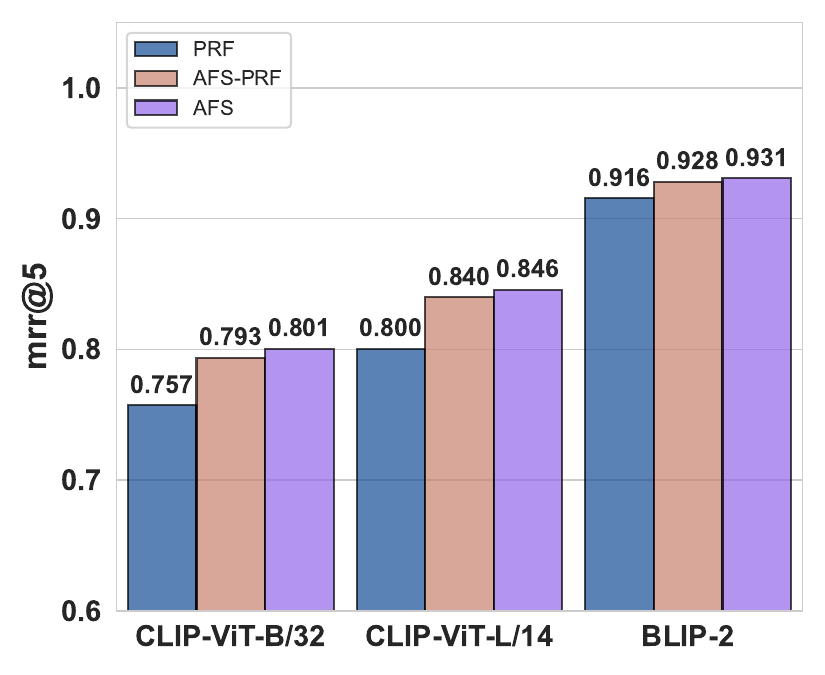}
        \\ (a) Flickr30K
        \label{fig:afs-prf-flickr}
    \end{minipage}
    \begin{minipage}{0.49\linewidth}
        \centering
        \includegraphics[width=\linewidth]{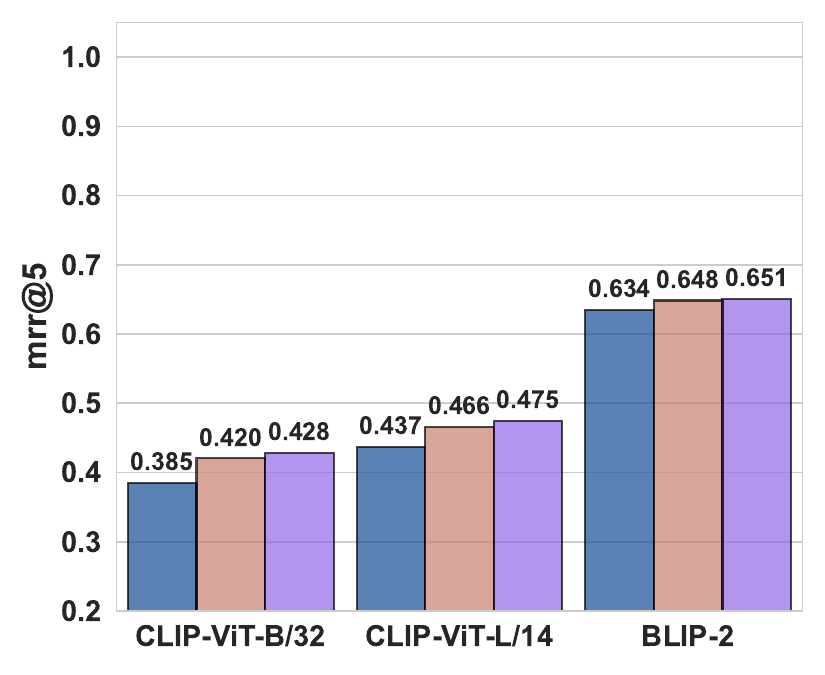}
        \\ (b) COCO
        \label{fig:afs-prf-coco}
    \end{minipage}
    
    \caption{\textbf{AFS-PRF Evaluation}. Comparison between AFS without generated captions (AFS-PRF), its full version, and Rocchio-based PRF.}
    \label{fig:afs-prf}
\end{figure}

Figure~\ref{fig:afs-prf} shows that AFS-PRF performs only slightly worse than the full AFS model while still outperforming PRF with Rocchio. This result suggests that AFS can serve as an effective tool for enabling pseudo-relevance feedback.

\subsection{Cross-attention Visualization Examples}
\label{sec:appendix_vis}

\begin{figure}[!h]
    \centering
    \includegraphics[width=0.9\linewidth]{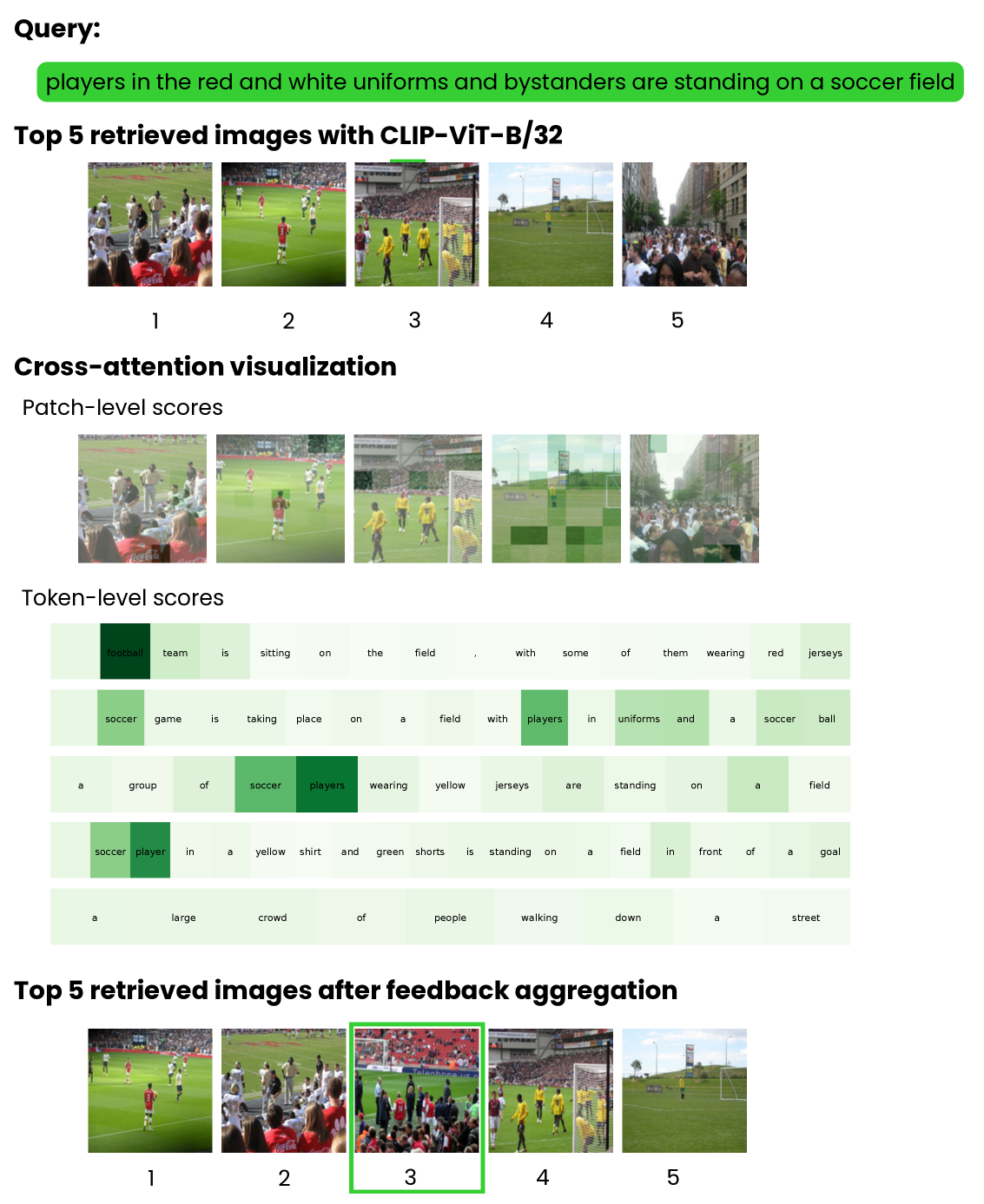}
    \caption{\textbf{Cross-attention visualization with CLIP-ViT-B/32}. This example shows a case when a ground-truth image was not used for relevance feedback aggregation, i.e., it was not initially among the top 5 retrieved images.}
    \label{fig:summarizer_attention_1}
\end{figure}

\begin{figure}[!t]
    \centering
    \includegraphics[width=0.9\linewidth]{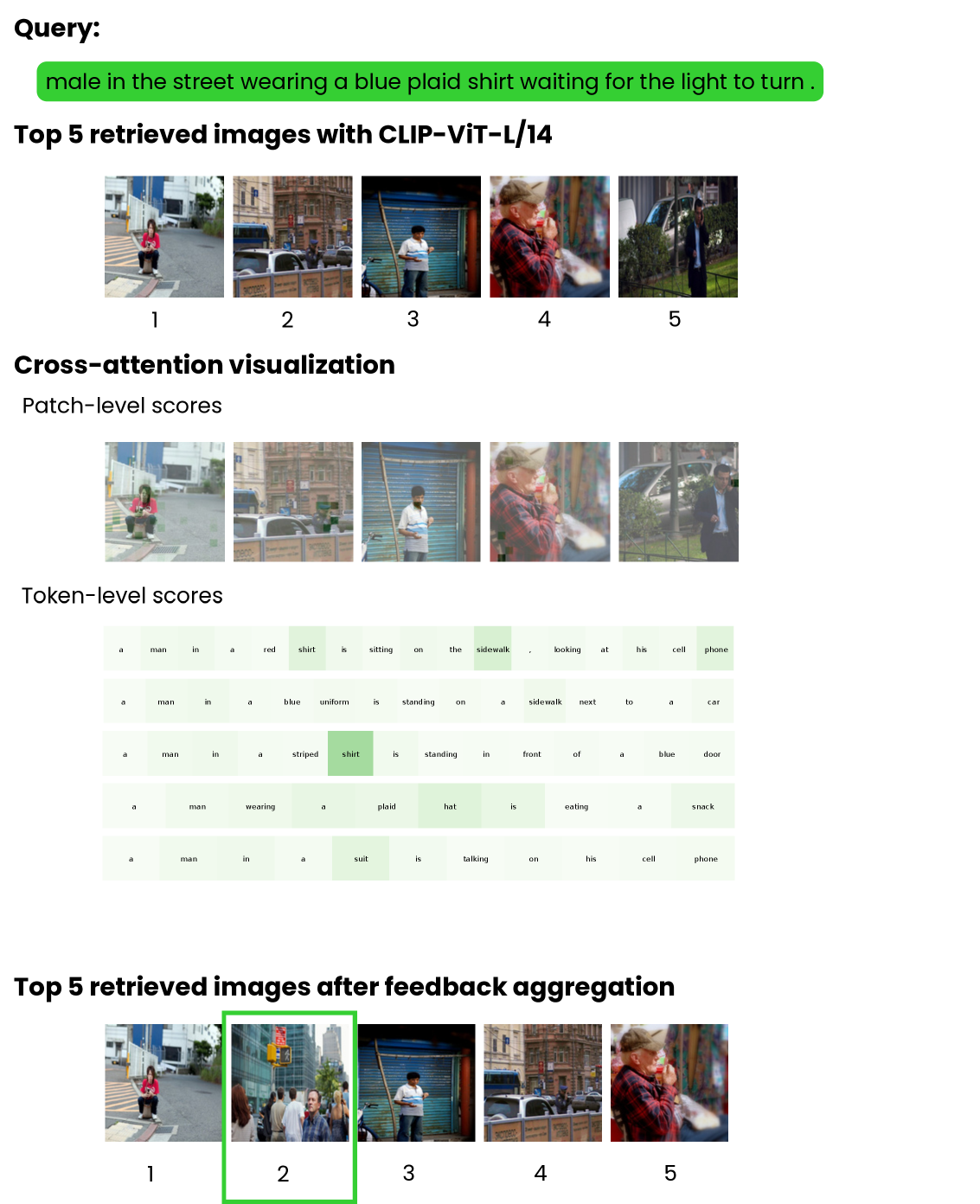}
    \caption{\textbf{Cross-attention visualization with CLIP-ViT-L/14}. The ground-truth image was not used for relevance feedback aggregation, i.e., it was not initially among the top 5 retrieved images.}
    \label{fig:summarizer_attention_2}
\end{figure}

\begin{figure}[!t]
    \centering
    \includegraphics[width=0.9\linewidth]{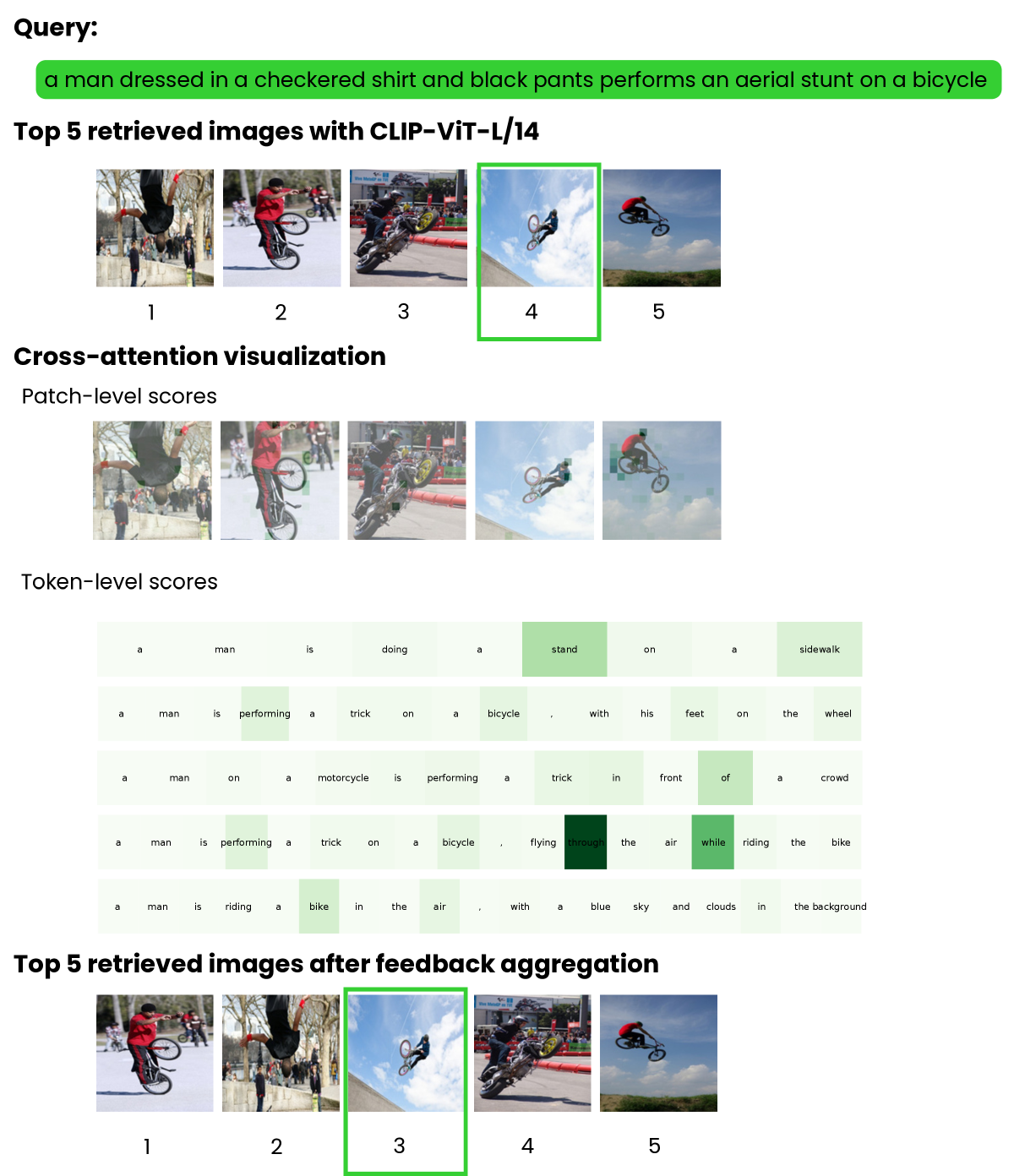}
    \caption{\textbf{Cross-attention visualization with CLIP-ViT-L/14.} AFS increases the rank of the ground-truth image.}
    \label{fig:summarizer_attention_3}
\end{figure}

\begin{figure}[!t]
    \centering
    \includegraphics[width=0.9\linewidth]{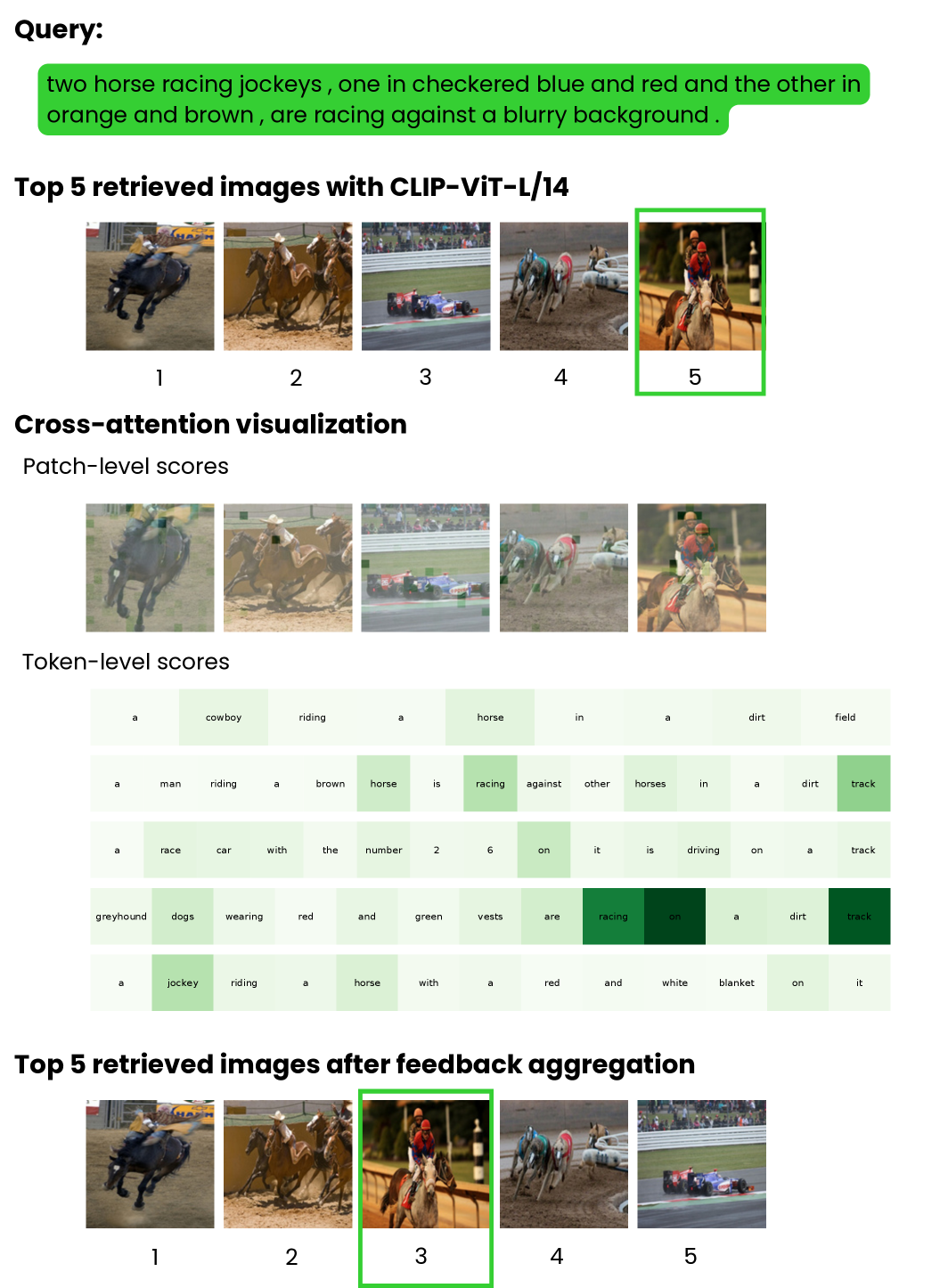}
    \caption{\textbf{Cross-attention visualization with CLIP-ViT-L/14}. AFS increases the rank of the ground-truth image.}
    \label{fig:summarizer_attention_4}
\end{figure}

We provide more cross-attention visualizations obtained from the AFS model, supplementing Section \ref{sec:results_explore_afs}. In Figures~\ref{fig:summarizer_attention_1} and~\ref{fig:summarizer_attention_2}, saliency maps are obtained for CLIP-ViT-B/32 and CLIP-ViT-L/14, respectively. In both cases, the ground-truth images were not among the top-5 initially retrieved items and, hence, were excluded from relevance feedback aggregation. Nevertheless, the obtained attention maps highlight semantically meaningful regions, and the refined query representations lead to better retrieval performance. Furthermore, Figures~\ref{fig:summarizer_attention_3} and~\ref{fig:summarizer_attention_4}, both using CLIP-ViT-L/14, show scenarios where the ground-truth image ranks improved after applying AFS. 

\section{Multi-turn Retrieval Results}
\label{sec:appendix_multiturn}

\begin{figure}[!t]
    \centering
    
    \begin{minipage}{0.32\linewidth}
        \centering
        \includegraphics[width=\linewidth]{figures/multi-turn/flickr_clip-vit-base-patch32_mrr5.pdf}
        \\ (a) CLIP-ViT-B/32
        \label{fig:multiturn_clip_base}
    \end{minipage}
    \begin{minipage}{0.32\linewidth}
        \centering
        \includegraphics[width=\linewidth]{figures/multi-turn/flickr_clip-vit-large-patch14_mrr5.pdf}
        \\ (b) CLIP-ViT-L/14
        \label{fig:multiturn_clip_large}
    \end{minipage}
    \begin{minipage}{0.32\linewidth}
        \centering
        \includegraphics[width=\linewidth]{figures/multi-turn/flickr_blip2-itm-vit-g_mrr5.pdf}
        \\ (c) BLIP-2
        \label{fig:fig:multiturn_blip2}
    \end{minipage}
    
    \begin{minipage}{0.32\linewidth}
        \centering
        \includegraphics[width=\linewidth]{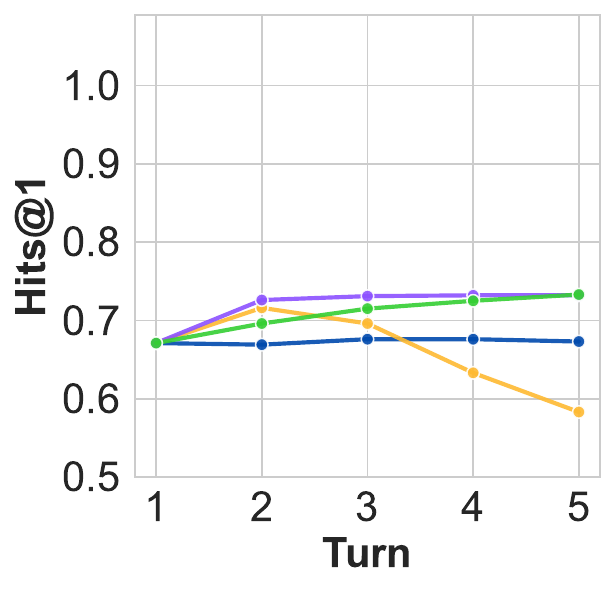}
        \\ (d) CLIP-ViT-B/32
        \label{fig:multiturn_clip_base}
    \end{minipage}
    \begin{minipage}{0.32\linewidth}
        \centering
        \includegraphics[width=\linewidth]{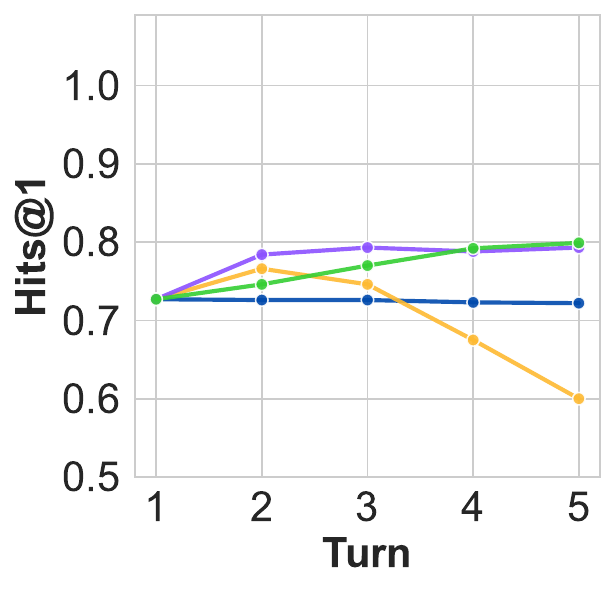}
        \\ (e) CLIP-ViT-L/14
        \label{fig:multiturn_clip_large}
    \end{minipage}
    \begin{minipage}{0.32\linewidth}
        \centering
        \includegraphics[width=\linewidth]{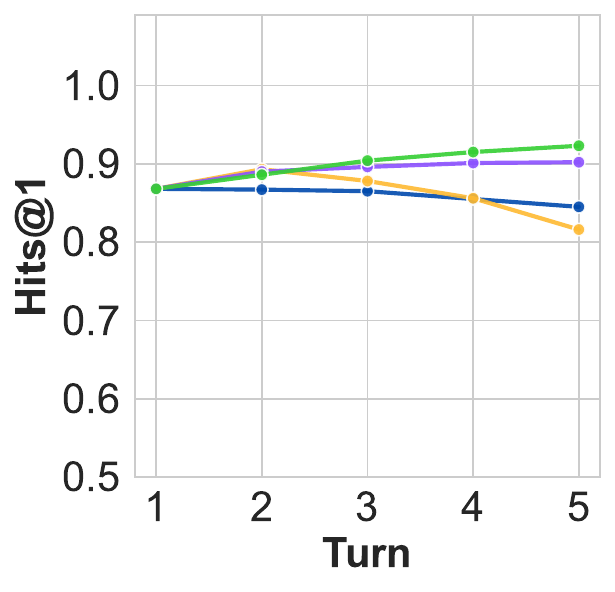}
        \\ (f) BLIP-2
        \label{fig:fig:multiturn_blip2}
    \end{minipage}

    \begin{minipage}{0.32\linewidth}
        \centering
        \includegraphics[width=\linewidth]{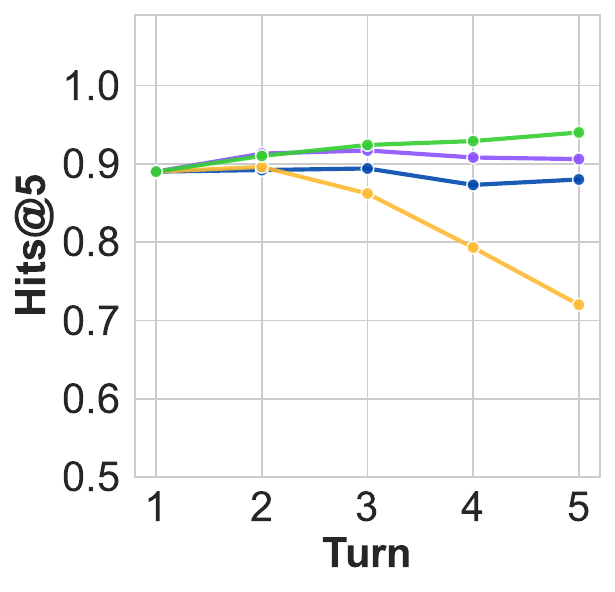}
        \\ (g) CLIP-ViT-B/32
        \label{fig:multiturn_clip_base}
    \end{minipage}
    \begin{minipage}{0.32\linewidth}
        \centering
        \includegraphics[width=\linewidth]{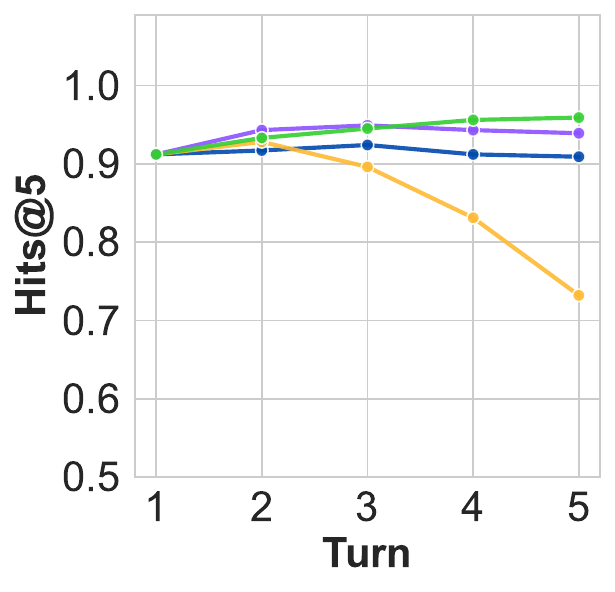}
        \\ (h) CLIP-ViT-L/14
        \label{fig:multiturn_clip_large}
    \end{minipage}
    \begin{minipage}{0.32\linewidth}
        \centering
        \includegraphics[width=\linewidth]{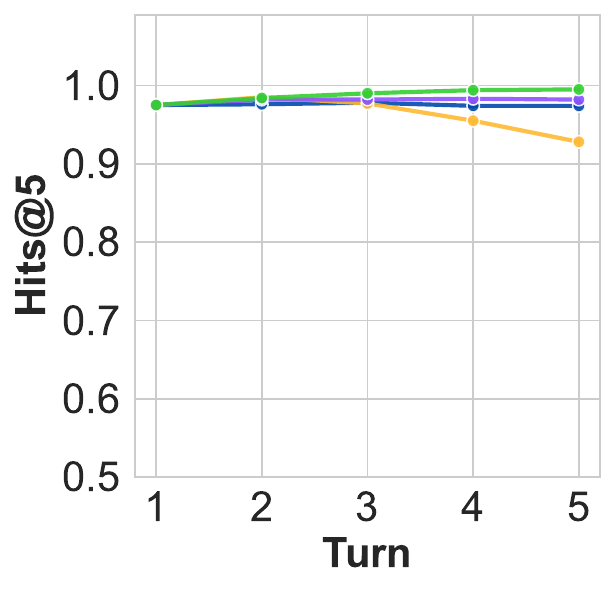}
        \\ (i) BLIP-2
        \label{fig:fig:multiturn_blip2}
    \end{minipage}
    
    \caption{Multi-turn retrieval performance with relevance feedback on Flickr30K.}
    \label{fig:multiturn_flickr_full}
\end{figure}

\begin{figure}[!h]
    \centering
    \begin{minipage}{0.32\linewidth}
        \centering
        \includegraphics[width=\linewidth]{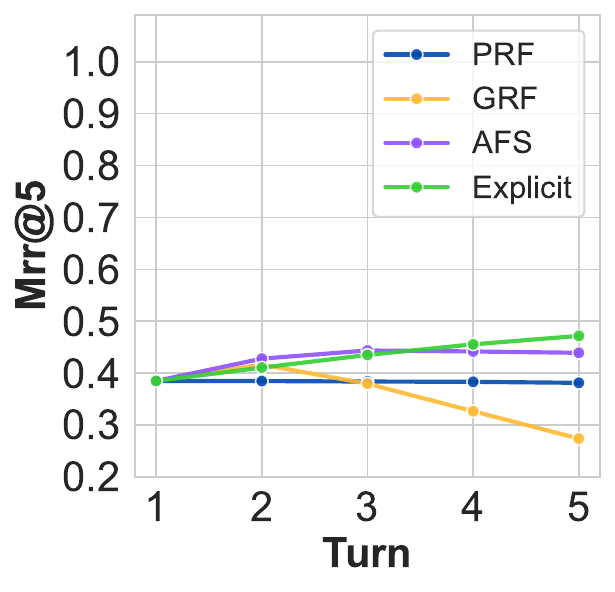}
        \\ (a) CLIP-ViT-B/32
        \label{fig:multiturn_clip_base}
    \end{minipage}
    \begin{minipage}{0.32\linewidth}
        \centering
        \includegraphics[width=\linewidth]{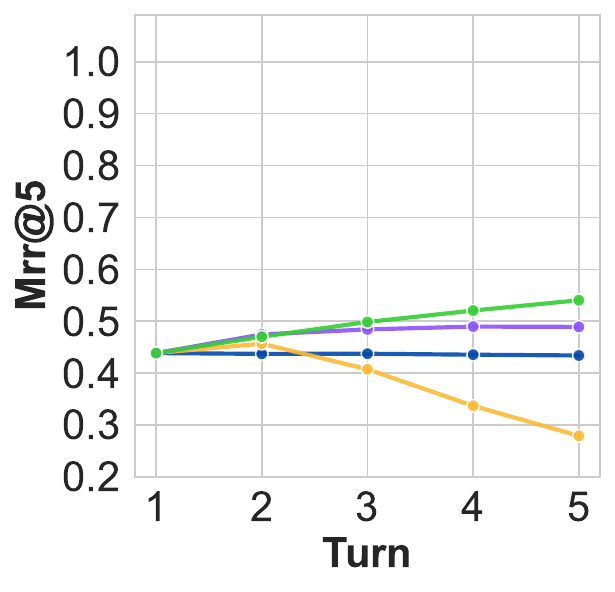}
        \\ (b) CLIP-ViT-L/14
        \label{fig:multiturn_clip_large}
    \end{minipage}
    \begin{minipage}{0.32\linewidth}
        \centering
        \includegraphics[width=\linewidth]{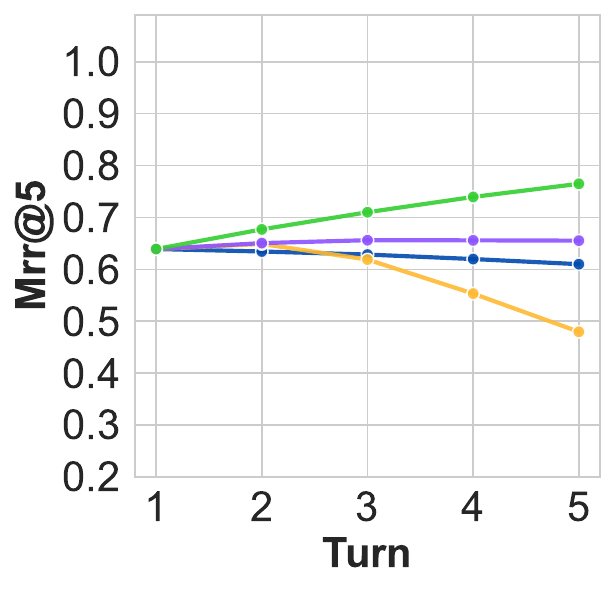}
        \\ (c) BLIP-2
        \label{fig:multiturn_blip2}
    \end{minipage}
    
    \begin{minipage}{0.32\linewidth}
        \centering
        \includegraphics[width=\linewidth]{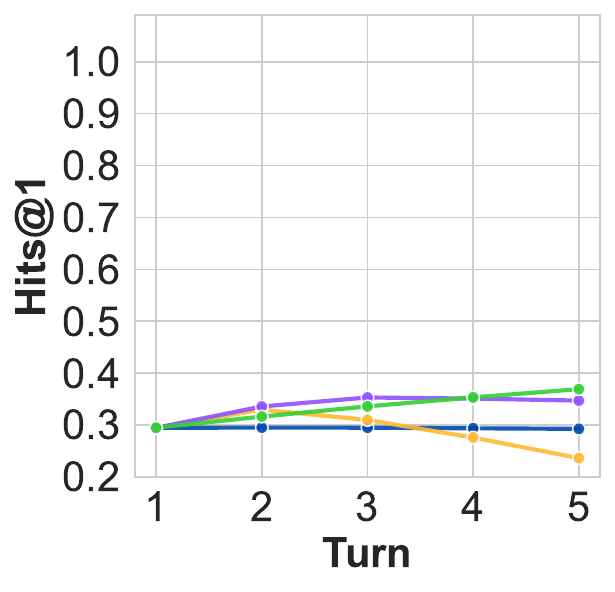}
        \\ (d) CLIP-ViT-B/32
        \label{fig:multiturn_clip_base}
    \end{minipage}
    \begin{minipage}{0.32\linewidth}
        \centering
        \includegraphics[width=\linewidth]{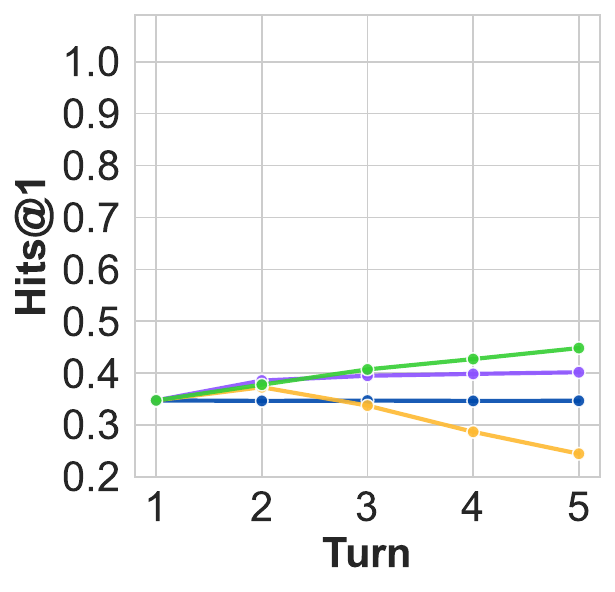}
        \\ (e) CLIP-ViT-L/14
        \label{fig:multiturn_clip_large}
    \end{minipage}
    \begin{minipage}{0.32\linewidth}
        \centering
        \includegraphics[width=\linewidth]{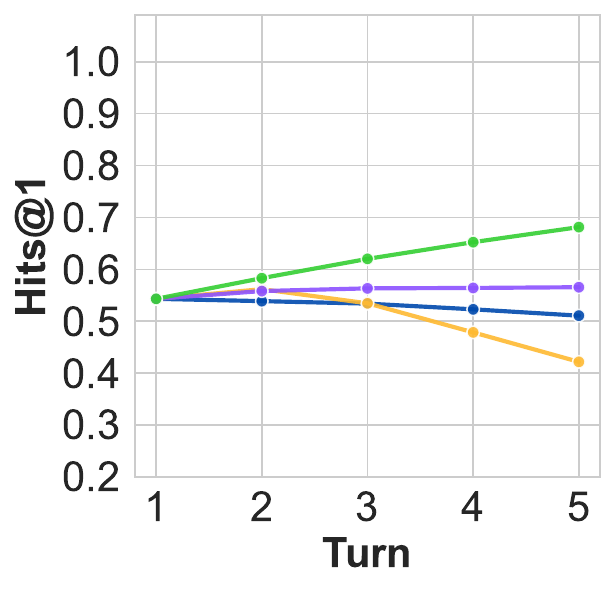}
        \\ (f) BLIP-2
        \label{fig:fig:multiturn_blip2}
    \end{minipage}

    \begin{minipage}{0.32\linewidth}
        \centering
        \includegraphics[width=\linewidth]{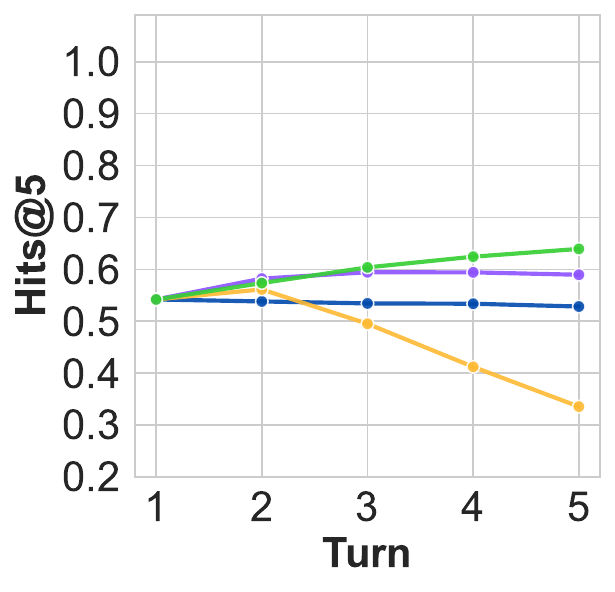}
        \\ (g) CLIP-ViT-B/32
        \label{fig:multiturn_clip_base}
    \end{minipage}
    \begin{minipage}{0.32\linewidth}
        \centering
        \includegraphics[width=\linewidth]{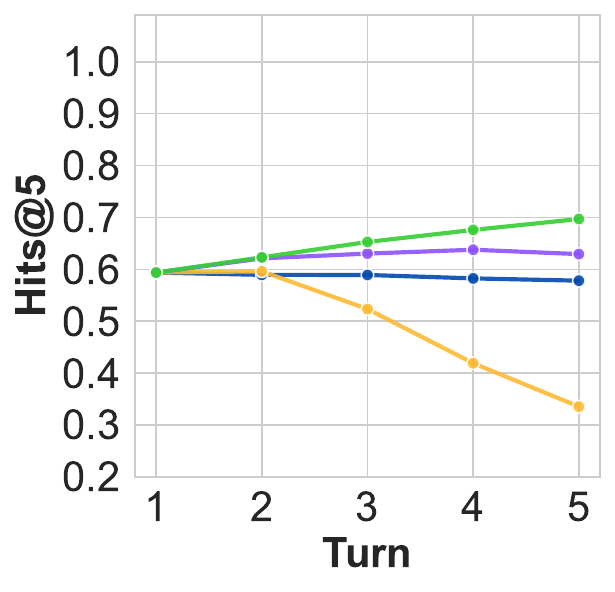}
        \\ (h) CLIP-ViT-L/14
        \label{fig:multiturn_clip_large}
    \end{minipage}
    \begin{minipage}{0.32\linewidth}
        \centering
        \includegraphics[width=\linewidth]{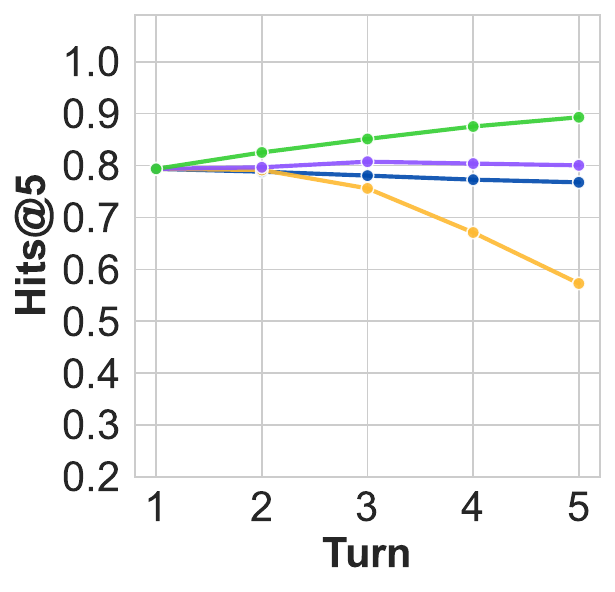}
        \\ (i) BLIP-2
        \label{fig:fig:multiturn_blip2}
    \end{minipage}
    
    \caption{Multi-turn retrieval performance with relevance feedback on COCO.}
    \label{fig:multiturn_coco_full}
\end{figure}

This section presents the extended results of the multi-turn retrieval with relevance feedback aggregation. Figures \ref{fig:multiturn_flickr_full} and \ref{fig:multiturn_coco_full} show the performance metrics achieved across retrieval rounds for the Flickr30K and COCO datasets. The obtained results complement the findings discussed in Section~\ref{sec:benchmarking_res}. Specifically, explicit feedback with ground truth captions continuously improves retrieval performance according to all metrics. Generative relevance feedback, however, leads to performance degradation starting from the third retrieval round. The attentive feedback summarizer demonstrates an increase in performance at round 2, gradually converging in rounds 3-5, avoiding query drift without ground-truth captions.

\section{Interactive Retrieval Demo}
\label{sec:appendix_demo}

In Section~\ref{sec:discussion}, we outlined future directions for combining relevance feedback techniques with user interactions to create more interactive retrieval systems. While some recent work explores chat-based interactions via LLMs \cite{lee2024interactive, zhu2024enhancing}, we investigate an alternative approach based on direct visual interaction with images. Specifically, we developed a prototype interface where users can retrieve images based on a textual query. The interface, then, allows users to annotate retrieved image regions by drawing bounding boxes indicating relevance or irrelevance to their search intent (Figure~\ref{fig:demo_fig}). 

We explored how the bounding boxes could be integrated into the inference process of the AFS model working without synthetic captions (Appendix \ref{sec:appendix_afs-prf}). Specifically, we modified the cross-attention scores based on a simple heuristic: attention weights were increased for image patches corresponding to regions marked as relevant by the user and decreased for those identified as irrelevant. The magnitude of the increase is a hyperparameter and can be defined through configurations. These adjustments were applied at inference time without requiring any additional training or fine-tuning of the model.

As shown in Figure~\ref{fig:demo_fig}, this approach effectively shifts the model's attention toward user-indicated regions, enabling more targeted retrieval responses. While our method is based on fixed weight adjustments, it demonstrates how user input can guide attention in relevance-based models. In the future, more sophisticated techniques could be explored. For instance, it is possible to add learnable token-level embeddings based on user feedback. This would require training the model with region-level relevance annotations, which are currently not available.

The code for the prototype is available at: \url{https://github.com/bulatkh/visualref/tree/wacv_demo}

\begin{figure}[!h]
    \centering
    \begin{subfigure}[b]{0.99\linewidth}
        \includegraphics[width=\linewidth]{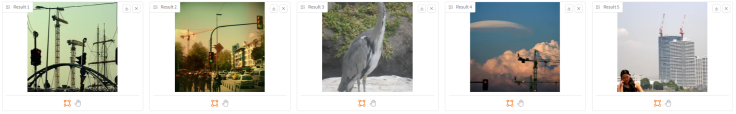}
        \caption{Initial Retrieval Results}
        \label{fig:demo_initial}
    \end{subfigure}
    
    \begin{subfigure}[b]{0.99\linewidth}
        \includegraphics[width=\linewidth]{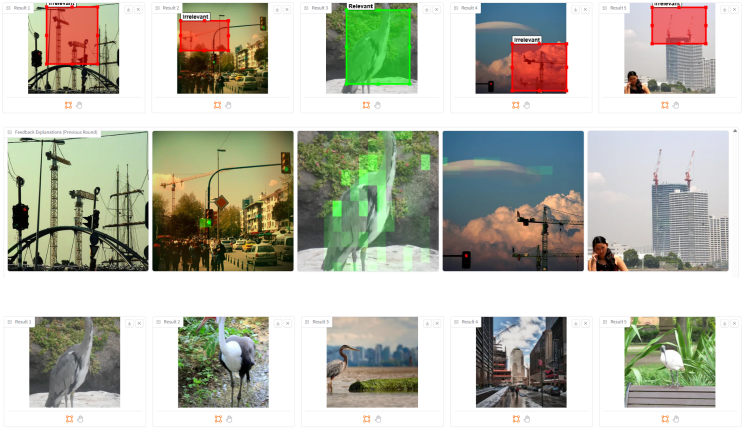}
        \caption{Focus on birds}
        \label{fig:demo_birds}
    \end{subfigure}
    
    \begin{subfigure}[b]{0.99\linewidth}
        \includegraphics[width=\linewidth]{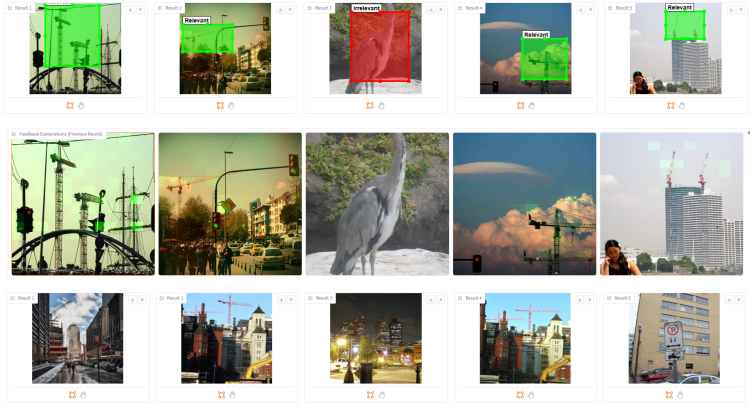}
        \caption{Focus on machines}
        \label{fig:demo_machine}
    \end{subfigure}
    \caption{\textbf{Interactive Retrieval Demo.} (a) Initial retrieval results for the textual query \textit{"Crane in the city"}. (b) The user specifies their intent by marking the bird as relevant and industrial cranes as irrelevant. (c) The user instead focuses on industrial cranes. In both (b) and (c), the first row shows examples of manual relevance annotations, the second row visualizes cross-attention scores after applying the feedback, and the third row presents the updated retrieval results using AFS with user signals.}
    \label{fig:demo_fig}
\end{figure}

\section{Limitations} 
\label{sec:appendix_limitations}

Our study systematically evaluated relevance feedback strategies with pre-trained VLM backbones. However, we do not compare against alternative query adjustment methods, such as query rewriting or prompt engineering with LLMs-in-the-loop. Future work can address this gap and explore the interplay between representation-level refinement and natural language query adjustments. This direction is particularly relevant for dialogue-aware image retrieval, where systems must model contextual coherence across multiple modalities and turns, resolve ambiguous references, and encode rich dialogue context for effective prompting of VLMs, which often struggle with non-descriptive queries~\cite{rodriguez2024does}. 

Further, GRF and AFS strategies introduce some additional overhead. GRF requires running an image captioning model in the background to generate captions for candidate images. AFS, on the other hand, operates at runtime, although its size remains modest (under 20 million parameters), especially compared to LLMs, which can contain billions of parameters. Therefore, we suggest that developers carefully weigh the trade-off between the retrieval gains offered by relevance feedback and the computational (and environmental) costs associated with deploying these methods.

\revision{Finally, we tested the proposed relevance feedback strategies on general-purpose image retrieval datasets. In future work, we are planning to evaluate these methods in domain-specific retrieval tasks.}

\end{document}